\documentclass[journal]{paper}
\usepackage[utf8]{inputenc}
\usepackage{hyperref}       
\usepackage{graphicx} 
\usepackage{caption} 
\usepackage{multirow}
\usepackage{subcaption}
\usepackage{color}
\usepackage{amsmath}
\usepackage{bm}
\usepackage{amsfonts}
\usepackage{makecell}
\usepackage{xcolor} 

\begin{document}

\title{Causal Contextual Prediction for Learned Image Compression}

 \author{Zongyu Guo, Zhizheng Zhang, Runsen Feng and
	Zhibo Chen,~\IEEEmembership{Senior Member,~IEEE,}
\thanks{
Zongyu Guo, Zhizheng Zhang, Runsen Feng and Zhibo Chen are with the Department of Electronic Engineer and Information Science, University of Science and Technology of China, Hefei, Anhui, 230026, China. Corresponding Author: Zhibo Chen (chenzhibo@ustc.edu.cn).

This work was supported in part by NSFC under Grant U1908209, 61632001 and the National Key Research and Development Program of China 2018AAA0101400.
}
}

\markboth{IEEE Transactions on Circuits and Systems on Video Technology}%
{Shell \MakeLowercase{\textit{et al.}}: Bare Demo of IEEEtran.cls for IEEE Journals}

\maketitle

\begin{abstract}
\inputencoding{utf8}
Over the past several years, we have witnessed impressive progress in the field of learned image compression. Recent learned image codecs are commonly based on autoencoders, that first encode an image into low-dimensional latent representations and then decode them for reconstruction purposes. To capture spatial dependencies in the latent space, prior works exploit hyperprior and spatial context model to build an entropy model, which estimates the bit-rate for end-to-end rate-distortion optimization. However, such an entropy model is suboptimal from two aspects: (1) It fails to capture global-scope spatial correlations among the latents. (2) Cross-channel relationships of the latents remain unexplored. In this paper, we propose the concept of separate entropy coding to leverage a serial decoding process for causal contextual entropy prediction in the latent space. A \textit{causal context model} is proposed that separates the latents across channels and makes use of channel-wise relationships to generate highly informative adjacent contexts. Furthermore, we propose a \textit{causal global prediction model} to find global reference points for accurate predictions of undecoded points. Both these two models facilitate entropy estimation without the transmission of overhead. In addition, we further adopt a new group-separated attention module to build more powerful transform networks. Experimental results demonstrate that our full image compression model outperforms standard VVC/H.266 codec on Kodak dataset in terms of both PSNR and MS-SSIM, yielding the state-of-the-art rate-distortion performance.

\end{abstract}

\begin{IEEEkeywords}
Learned image compression, causal context model, causal global prediction, improved entropy model.
\end{IEEEkeywords}

\IEEEpeerreviewmaketitle

\section{Introduction}

\IEEEPARstart{L}{ossy} image compression is a fundamental technique for image transmission and storage. Since the concept of hybrid coding was proposed \cite{habibi1974hybrid,Forchheimer1981differential}, the hybrid coding framework has shown strong vitality in the field of media compression. This coding framework, combining prediction and transformation, has been improved for decades, not only to achieve better performance but also to meet the rising demand of novel multimedia applications.
Since the standardization of H.264/AVC in 2003 \cite{wiegand2003overview}, the technique of intra prediction has become an important component of image compression. 
Researchers have been developing increasingly complex intra prediction methods for decades, expecting more accurate predictions and thus more efficient compression. Intra prediction basically tries to capture the spatial correlations of the image to reduce the spatial redundancies in the compressed bitstream.

Over the past few years, the progress of learning-based image compression is impressive \cite{toderici2017full,balle2018variational,agustsson2019generative,cheng2020learned}. Despite the short history of this field, end-to-end optimized compression models have shown the potential to outperform traditional manually designed codecs. 
Learned image compression commonly follows a pipeline consisting of transformation, quantization and lossless entropy coding, which resembles traditional transform coding \cite{goyal2001theoretical}. Specifically, a non-linear transform is first performed on the input image, mapping it to latent representations. Then the latents are quantized, yielding discrete representations. 
To losslessly compress the discrete latents, an entropy model is used to estimate their discrete entropy \cite{balle2017end}; this model is later improved as a hierarchical entropy model, \textit{i.e.}, a hyperprior model \cite{balle2018variational}. More recently, several studies introduce the concept of context model \cite{minnen2018joint,lee2019context}, which is an autoregressive model over latents. Such a context model usually employs mask convolution \cite{van2016pixel} to aggregate local contexts for efficient entropy coding.

Worthy of mention is that the non-linear transform and entropy model play different roles for compression although they both can decorrelate natural images.
While the non-linear transform disentangles the image into compact latent representations, the entropy model exploits a probabilistic structure to establish an accurate estimation for bit-rate and eventually leads to the generation of smaller files \cite{balle2018variational,minnen2018joint,lee2019context}.
As a part of the entropy model, the autoregressive context model \cite{minnen2018joint,lee2019context} is more like a \textit{prediction} module, which leverages adjacent available latents to predict undecoded points.
Such an autoregressive model incurs a significant computational penalty during decoding due to the serial decoding process, but it largely improves
the rate-distortion performance.

\begin{figure}[t]
    \centering
    \includegraphics[scale=0.84, clip, trim=18.4cm 10.6cm 1.4cm 2.2cm]{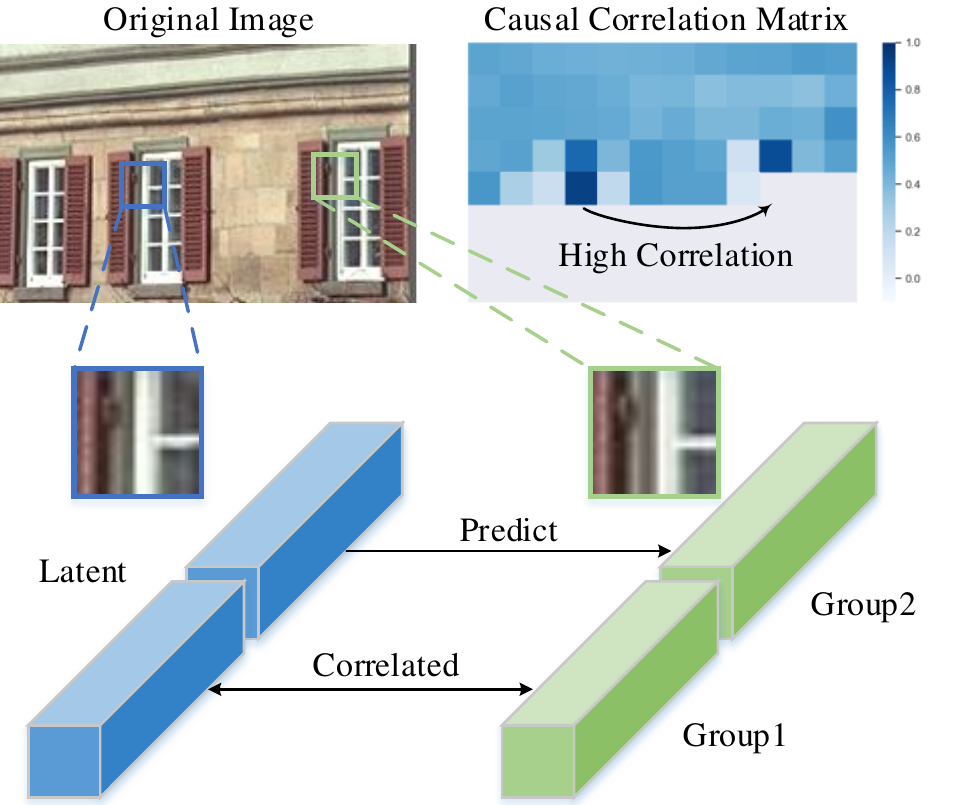}
    \caption{An overview of our proposed causal global prediction model. A correlation matrix is calculated from the first channel group to describe global dependencies. This correlation matrix is \textit{causal} and guides the \textit{global prediction} for these undecoded channels (the second channel group).}
\label{figure1}
\end{figure}

However, the entropy model, which incorporates adjacent contexts and hierarchical priors, is suboptimal in terms of two aspects.
On the one hand, adjacent context modeling is not allowed to capture global correlations of the latents which are beneficial for sufficient information decorrelation. 
Some previous works \cite{li2018learning,johnston2018improved} present spatial-adaptive encoders/decoders, the transformation of which is adaptive to global image contents. Despite this, the exploitation of global contexts within the entropy model is still ignored.
In theory, a pair of image patches with similar context should be transformed into similar latent representations, even if they are spatially far away from each other (see Fig.\ref{figure1}).
Thus, going beyond the context modeling of adjacent contexts towards a global-scope context modeling is a theoretically powerful modification to deliver more accurate entropy estimation.
However, a critical challenge is imposed when performing global-scope context modeling in the entropy model. That is,
\textit{how can global correlation information be established at the decoder?}
We experimentally demonstrate that it is not valuable to explicitly describe global references through the transmission of side information (see Section \ref{section3b} for details), which implies the demand of a special design to realize global prediction.
On the other hand, in addition to spatial redundancies, cross-channel relationships remain unexplored. In fact, after learning-based transformation, information located spatially is embedded into channel dimension. Therefore, cross-channel relationship modeling is meaningful as well.
Compared with traditional compression, this is a special issue for neural image compression because the latent space is now 3-D. 
An optimal entropy model should be able to eliminate channel-wise redundancies for improved entropy estimation. 

In this paper, to address the two abovementioned issues, we propose a causal context model and a causal global prediction model for more accurate entropy estimation. Experimentally, we find that the spatial correlations of the entire latents can be approximated by partial channels, indicating that there are redundancies among different channels.
We thus design a causal context model with an improved mask convolution, where we separate the latents into two groups. 
Subsequently, we propose a novel causal global prediction model to reduce global redundancies among the latent variables. Specifically, once the first latent group is decoded (Group 1 in Fig.\ref{figure1}), the proposed causal context model will aggregate both the spatially adjacent latents and the first half-channel elements in the current spatial location to yield highly informative adjacent contexts. In addition. the proposed causal global prediction model will leverage the decoded group to model global correlations, establishing a causal correlation matrix that guides the global predictions for the remaining undecoded channels (Group 2 in Fig.\ref{figure1}).
Different from previous hyperprior model, the proposed causal global prediction model does not require the transmission of overhead to describe global correlations. 

Through causal contextual prediction within the entropy model, the contexts are embedded into two separate channel groups, where one latent group is directly inferred as usual while the other group is causally inferred upon the condition of the prior one. 
In addition, we adopt a new group-separated attention module to strengthen the representation ability of the transform networks. Motivated by \cite{zhang2020multi,zhang2020resnest}, this module enables independent feature-map attention across separated groups and is demonstrated to be more powerful than the attention mechanism used in previous compression method \cite{cheng2020learned}.
Our technical contributions can be summarized as follows:

\begin{itemize}
    \item We propose the concept of separate entropy coding by dividing the latent representation into two channel groups for more effective context modeling. We thereby propose a causal context model that makes use of cross-channel redundancies to generate highly informative adjacent contexts.
    \item We pinpoint that exploiting global-scope contexts is vital for the entropy modeling, and propose a causal global prediction model to conduct prediction in a global scope. By extending the concept of separate entropy coding, this model does not require extra transmission of overhead but can still establish global reference information.
    \item We adopt group-separated attention module to strengthen the non-linear transform networks, which is demonstrated to be more powerful than previous attention designs. We take this point as a side contribution in this paper.
\end{itemize}



We integrate the proposed causal context model and causal global prediction model, along with the group-separated attention module, building our learned image compression network. We verify the effectiveness of each component with sufficient ablation studies. Experimental results demonstrate that our approach outperforms previous learned image compression schemes \cite{minnen2018joint,cheng2020learned} in terms of both PSNR and MS-SSIM \cite{wang2004image}. Moreover, compared with traditional codec VTM 8.0\footnote{\url{https://vcgit.hhi.fraunhofer.de/jvet/VVCSoftware_VTM/-/tree/VTM-8.0}}, our method achieves 5.1\% BD rate savings on the Kodak dataset\footnote{\url{http://r0k.us/graphics/kodak/}} in terms of PSNR. 

Our work described in this paper is an extension of our pioneering short conference paper \cite{guo20203}. In \cite{guo20203}, we introduce the causal context model\footnote{In \cite{guo20203}, it is termed 3-D context entropy model. We here term it as causal context model to avoid ambiguity with 3-D mask convolution.}, which is the first instantiation of the concept of separate entropy coding. The differences between this paper and \cite{guo20203} are described as follows. Firstly, here we extend the concept of separate entropy coding to propose a novel causal global prediction model that successfully reduces global redundancies of the latents and achieves more efficient entropy estimation. Secondly, we provide detailed analyses to explain the design of separate entropy coding, as illustrated in Section \ref{section4d}. Thirdly, an improved attention module is employed to enhance the non-linear transform networks. Fourthly, we conduct extensive ablation studies that verify the effectiveness of our proposed new techniques. We also evaluate the coding time of our method.

The remainder of this paper is organized as follows. Section \ref{section2} briefly overviews some relevant approaches. Section \ref{section3} illustrates the motivation of global prediction in the entropy model. Section \ref{section4} introduces our proposed causal context model and causal global prediction model along with the proposed group-separated attention layer. Section \ref{section5} is the experimental section. We conclude this paper in the last section.

\section{Related Work\label{section2}}

\subsection{Traditional Image Compression}
Traditional image compression standards, such as JPEG \cite{wallace1992jpeg}, JPEG2000 \cite{rabbani2002jpeg2000} and BPG (HEVC intra) \footnote{\url{https://bellard.org/bpg/}}, are widely used in practice. They heavily rely on manually designed modules but always follow the concept of transform coding \cite{goyal2001theoretical}. Since the standardization of H.264/AVC \cite{wiegand2003overview}, intra prediction has become an important component of image compression. In the case of H.264, there are a total of 9 intra prediction modes. With regard to H.265/HEVC \cite{sullivan2012overview}, 35 optional intra prediction modes are adopted for different sizes of prediction units (PUs). The ongoing H.266/VVC \cite{vvccite} further applies 65 intra modes. Intra prediction plays an important role in reducing the spatial redundancies of images in hybrid compression framework.

Additionally, it is worthwhile to note the intra block prediction (IBC) technique, which was first standardized in the screen content coding extensions \cite{xu2015overview} of HEVC. Intra block compensation is observed to be effective for screen contents. Therefore, IBC utilizes a bit-consuming displacement vector (referred as block vector or BV) to represent the relative displacement between the current prediction unit and the reference block. Such classical technique can conduct global searching for more efficient intra prediction and it is similar to the global prediction idea in this paper. However, our solution for global prediction of the latents does not require the transmission of prediction vectors.

\subsection{Learned Image Compression}

\subsubsection{Framework} In the field of learned image compression, researchers initially study the problem of non-differentiable quantization in artificial neural network \cite{theis2017lossy,agustsson2017soft}. The first learned compression method to outperform JPEG is an RNN-based model \cite{toderici2017full}. This framework also supports coding scalability \cite{li2001overview,zhang2019learned} but is only optimized for distortion. The work of \cite{balle2017end} attempts to estimate the \textit{rate} as discrete entropy. These works make it feasible to train an end-to-end compression network optimized for the rate-distortion trade-off. In \cite{balle2018variational}, a hyperprior model is proposed that transmits extra side information to model the spatial structure of the latents. This work also parameterizes the distribution of the latents to a zero-mean Gaussian scale model (GSM). After that, the parametric form of latent distribution is improved from Single Gaussian Model (SGM) \cite{minnen2018joint} to Gaussian Mixture Model (GMM) \cite{cheng2020learned}. In addition, motivated by PixelCNN \cite{van2016pixel}, the concept of context model is introduced for efficient entropy estimation \cite{minnen2018joint,lee2019context}. However, this type of model only covers local regions but does not pay attention to global scope.

\subsubsection{Global Context}
An early work \cite{johnston2018improved} uses a spatially adaptive post-process to dynamically adjust bit rate according to a target reconstruction quality. Later, both \cite{li2018learning} and \cite{mentzer2018conditional} use an importance map to for more adaptive bit rate allocation. All of those works \cite{johnston2018improved,li2018learning,mentzer2018conditional} incorporate image content to affect compression, but they do not have explicit entropy model. Recently, Lee et al. \cite{lee2019hybrid} also explore to utilize global context within the entropy model. However, they are unable to establish accurate global references at the decoder side. Aggregating possible global contexts is hard to determine accurate global reference information, thereby leading to the suboptimal performance of their entropy model. 

\subsubsection{Cross-Channel Relationship}
Previous works \cite{mentzer2018conditional,chen2019neural,li2020learning} also consider the relationships between latent channels. One early work \cite{mentzer2018conditional} does not have a parametric entropy model which limits their performance.
Unlike separate entropy coding in our paper, Chen et al. \cite{chen2019neural} adopt a channel-autoregressive 3-D mask convolution that slides over the latent representations across channels. In their method, every latent element is conditioned on adjacent decoded elements that are spatial-channel neighbors. Therefore, when applying a $5\times5\times5$ mask convolution, they could only learn correlations from adjacent two channels. Li et al. \cite{li2020learning} propose a similar idea to ours but their entropy model is also autoregressive along channels, which complicates the practical entropy coding and thus increases the time complexity. Instead of channel-autoregressive entropy coding, here we demonstrate that separating the latents into two groups can help network adaptively learn appropriate context, which is simple but effective.

\section{Problem Definition\label{section3}}

\subsection{Learned Compression Framework}

In the framework of learned image compression, a natural image, denoted by \bm{$x$}, is first encoded as latent representations \bm{$y$} through an analysis transform $g_a(\bm{x}|\bm{\phi})$. Then the latents \bm{$y$} are quantized to discrete values \bm{$\hat{y}$}, which will be losslessly coded by algorithms such as arithmetic coding. At the decoder side, a synthesis transform $g_s(\bm{\hat{y}} | \bm{\theta})$ recovers $\bm{\hat{y}}$ to reconstruct an image \bm{$\hat{x}$}.
\begin{equation}
\begin{aligned}
\bm{y} & = g_a(\bm{x}|\bm{\phi}), \\
\bm{\hat{y}} &  = Q(\bm{y}), \\
\bm{\hat{x}} &  = g_s(\bm{\hat{y}}| \bm{\theta}).
\end{aligned}
\label{equ1}
\end{equation}
If taking entropy model into account, the above neural compression schemes can be intepreted as variational autoencoders (VAEs) \cite{kingma2013auto}. In particular, the rate-distortion objective of end-to-end compression has a strong relationship with the loss function of $\beta$-VAE \cite{higgins2016beta}, where there is a hyperparameter balancing the latent channel capacity with reconstruction quality:
\begin{equation}
\begin{aligned}
\mathcal{L} =\mathbb{E}_{\bm{x}\sim p_{\bm{x}}}[-\log_2 p_{\bm{\hat{y}}}(\bm{\hat{y}})] + \lambda \cdot \mathbb{E}_{\bm{x}\sim p_{\bm{x}}} [d(\bm{x}, \bm{\hat{x}})].
\end{aligned}
\label{equ2}
\end{equation}
The first term is the rate that corresponds to the cross entropy between the natural (marginal) distribution and the learned entropy model. The second term measures the reconstruction quality according to the given distortion metric $d$ (e.g., PSNR or MS-SSIM).
Imposing a variant constraint by adjusting $\lambda$ influences the disentangling effect of the non-lienar transform \cite{higgins2016beta} and determines the model rate. 
In the work of \cite{balle2018variational}, to reduce the spatial redundancies among latent variables, a hyperprior model is proposed, which assigns a few extra bits as side information to transmit some spatial structure information. Such a hyperprior model helps to learn an accurate entropy model and thereby achieves a better estimation of $p_{\bm{\hat{y}}}(\bm{\hat{y}})$. This hyper model can be roughly divided into a hyper analysis transform $h_a(\bm{y}|\bm{\phi_h})$ and a synthesis transform $h_s(\bm{\hat{z}} | \bm{\theta_h})$ as
\begin{equation}
\begin{aligned}
\bm{z} & = h_a(\bm{y}|\bm{\phi_h}), \\
\bm{\hat{z}} &  = Q(\bm{z}), \\
P_{\hat{y}|\hat{z}}(\bm{\hat{y}} &|\bm{\hat{z}})  \leftarrow h_s(\bm{\hat{z}} | \bm{\theta_h}).
\end{aligned}
\label{equ3}
\end{equation}
However, such spatial structure information is inaccurate because the side information is transmitted after several downsampling layers, leading to geometric information loss. To capture the adjacent correlations of \bm{$\hat{y}$}, a context model $f_c$ \cite{minnen2018joint,lee2019context} is equipped as a component of hyperprior model. It is usually in the form of mask convolution with parameter \bm{$\theta_c$}. At the expense of decoding speed, we now decode all points serially, \textit{e.g.}, in raster scan order. Then, the hyperprior entropy model can be depicted as follows:
\begin{equation}
\begin{aligned}
& \bm{c_n} = f_c(\bm{\hat{y}_{i_1}},\bm{\hat{y}_{i_2}},...,\bm{\hat{y}_{i_k}}|\bm{\theta_c}), \\
& P_{\hat{y}|\hat{z}}(\bm{\hat{y}|\bm{\hat{z}}})  \leftarrow h_s(\bm{\hat{z}}, \bm{c_n} | \bm{\theta_h}),
\end{aligned}
\label{equ4}
\end{equation}
where $\bm{i_1, i_2, ..., i_k}$ indicate the indices of the subset of decoded points for aggregating adjacent context \bm{$c_n$}. Since at any point the decoder can only access the already decoded latents, the local context model only touches the left and top points which are adjacent to the current decoding point. 

\subsection{Global Prediction\label{section3b}}

Basically, the idea of global prediction aims at further leveraging all previously decoded latents to predict the current decoding point, and this can be formulated as
\begin{equation}
\begin{aligned}
& \bm{c_n} = f_{c}(\bm{\hat{y}_{1}},\bm{\hat{y}_{2}},...,\bm{\hat{y}_{n-1}}|\bm{\theta_c}), \\
& P_{\hat{y}|\hat{z}}(\bm{\hat{y}|\bm{\hat{z}}})  \leftarrow h_s(\bm{\hat{z}}, \bm{c_n} |  \bm{\theta_h}),
\end{aligned}
\label{equ5}
\end{equation}
where $\bm{\hat{y}_{1}},\bm{\hat{y}_{2}},...,\bm{\hat{y}_{n-1}}$ represent all points that are already decoded.
According to Fig.\ref{figure1}, we observe that global spatial redundancy remains in the latents.
We herein conduct an exploratory experiment to verify that such global redundancies heavily influence the performance of the entropy model.

A common way to describe global correlation is to calculate the similarity between all points. However, in the task of compression where decoding is serial, we are only interested in the similarities\footnote{Here, we compute the cosine distance to measure similarity.} between all previously decoded points and the current decoding point. Therefore,
we calculate \textit{causal} correlation matrices (see Fig.\ref{figure1}) that help us search global reference points. We should keep in mind that the correlation matrix is not directly available at the decoder side.
The decoder cannot compute the similarity between other points and the current decoding point before it decodes out the specific feature vector in the current decoding location. 

\begin{table}[t]
\centering
\renewcommand{\arraystretch}{1.2}
\resizebox{0.8\columnwidth}{!}{
\begin{tabular}{|c|c|c|c|}
    \hline
    Model & Anchor & Case 1 & Case 2 \\
    \hline
     Rate (bpp) & 0.1588 & 0.1435 & 0.1799    \\
    \hline
     PSNR (dB) & 28.53  & 28.54 & 28.56 \\
    \hline
\end{tabular}}

\caption{Making use of global reference information is advantageous for entropy estimation. In case 1, the transmission cost of global reference information is not considered and thus case 1 is not a practical codec. In case 2 we assign additional bits to transmit the coordinates of reference points in advance. }
\label{table1}
\end{table}

We start with an ablation experiment to study the effect of global context on entropy model, where the baseline network already adopts hierarchical priors and local context model \cite{minnen2018joint}. As shown in Table \ref{table1}, in both case 1 and case 2, the decoder is assumed to be able to make use of the coordinates of global reference points for more efficient entropy estimation. However, in case 1, we do not calculate the transmission cost of the global reference information, while in case 2, we assign additional bits to transmit the coordinates of reference points in advance\footnote{Since coordinates are integers as well, we use another hierarchical entropy model to estimate the rate of reference coordinates.}. 
During the training stage in case 2, we modify the loss function to include the transmission overhead. 
Comparing case 1 and case 2 with the baseline model, we can conclude that incorporating global references into entropy model significantly improves entropy estimation. Specifically, the required bit-rate decreases from 0.1588 to 0.1435 bit per pixel (bpp) while the reconstruction quality is maintained as we that can observe from case 1.

The above experiments demonstrate the advantages of incorporating accurate global context into the entropy model. However, it also reveals a critical problem: \textit{how to efficiently establish accurate global correlation information at the decoder side?} In this paper, we propose a causal global prediction model that does not require transmission of overhead but can still conduct global prediction. It is based on the causal context model, which aims at generating highly informative adjacent context via latent separation.

\section{Proposed Method\label{section4}}

\subsection{Causal Context Model}

\begin{figure}[t]
 \centering

 \begin{subfigure}{\columnwidth}
 \includegraphics[scale=0.46, clip, trim=17.2cm 32.5cm 24cm 20.2cm]{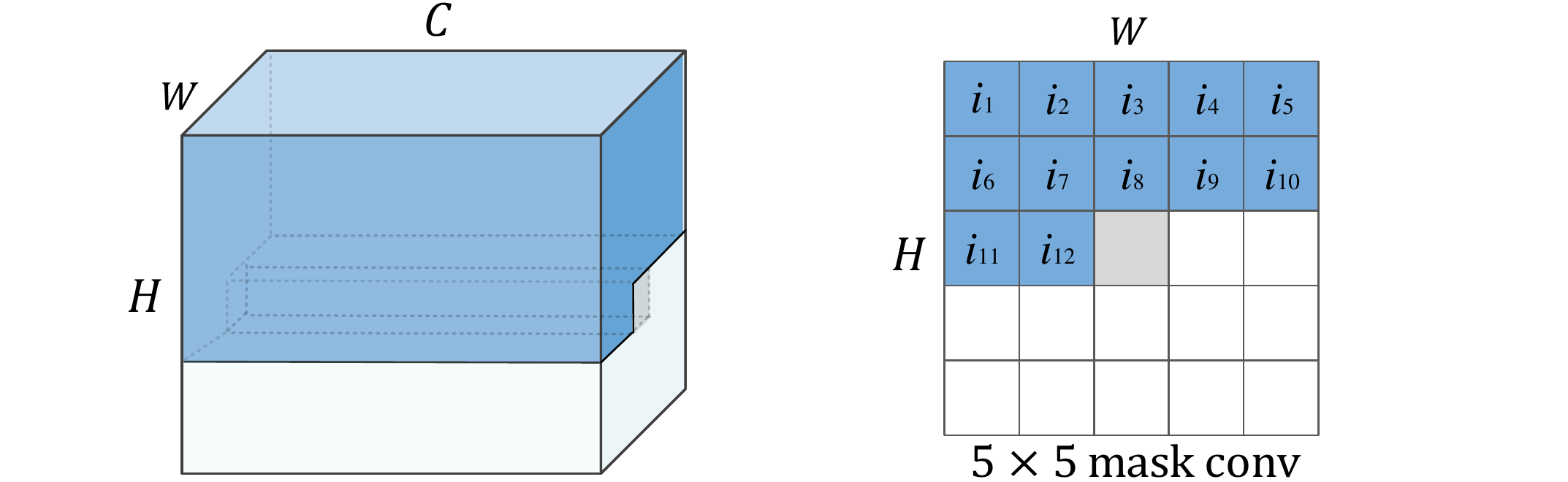}
 \caption{ \label{figure2a}}
 \end{subfigure}

 \begin{subfigure}{\columnwidth}
 \includegraphics[scale=0.46, clip, trim=17.2cm 24cm 24cm 27.5cm]{figures/context_model.pdf}
 \caption{ \label{figure2b}}
 \end{subfigure}

 \caption{(a) The structure of conventional mask convolution \cite{van2016pixel}. (b) The improved mask convolution in our proposed causal context model (denoted by Mask Conv +).}
\label{figure2}
\end{figure}

Inspired by the conditional RGB prediction of PixelCNN++ \cite{salimans2017pixelcnn++}, we propose a causal context model to improve entropy estimation, as described in our pioneer conference paper \cite{guo20203}. The causal context model separates the latents \bm{$\hat{y}_n$} in spatial location $n$ into two groups across channels. When it comes to decode \bm{$\hat{y}_n$}, the distribution of the first channel group \bm{$\hat{y}_{n,1}$} is predicted as usual, which adopts a mask convolution $f_{c,1}$ (as in Fig.\ref{figure2a}) to generate adjacent context $\bm{c_{n,1}}$ and combines with hyperpriors $\bm{\hat{z}}$. 
Thanks to the serial decoding process, once the first group \bm{$\hat{y}_{n,1}$} is decoded, we can take this known group to estimate the second group \bm{$\hat{y}_{n,2}$}. By modifying the conventional mask convolution, the improved mask convolution $f_{c,2}$, which is shown in Fig.\ref{figure2b}, is able to aggregate both the adjacent decoded latents and the first half latents in current spatial location, generating more informative adjacent contexts $\bm{c_{n,2}}$. The improved mask convolution can be regarded as a mutation between mask convolution and causal convolution \cite{oord2016wavenet}. It leverages the serial decoding process to generate causal contexts. 
The whole process can be formulated as follows:
\begin{equation}
\begin{aligned}
& \bm{c_{n,1}} = f_{c,1}(\bm{\hat{y}_{i_1}},\bm{\hat{y}_{i_2}},...,\bm{\hat{y}_{i_k}}), \\
& P_{\bm{\hat{y}_{n,1}|\hat{z}}}(\bm{\hat{y}_{n,1}|\bm{\hat{z}}})  \leftarrow h_{s,1}(\bm{\hat{z}},  \bm{c_{n,1}}| \bm{\theta_h}), \\
& \bm{c_{n,2}} = f_{c,2}(\bm{\hat{y}_{i_1}},\bm{\hat{y}_{i_2}},...,\bm{\hat{y}_{i_k}}, \bm{\hat{y}_{n,1}}), \\
& P_{\bm{\hat{y}_{n,2}|\hat{z}}}(\bm{\hat{y}_{n,2}|\bm{\hat{z}}})  \leftarrow h_{s,2}(\bm{\hat{z}}, \bm{c_{n,2}} | \bm{\theta_h}).
\end{aligned}
\label{equ6}
\end{equation}

This causal context model can effectively extract the cross-channel relationship to facilitate entropy estimation of the second channel group. Note that the concept of separate entropy coding actually divides the decoding of $\hat{y}_n$ into two stages. At different stages, the network parameters are independent. As a result, the following parameter estimation modules $h_{s,1}$ and $h_{s,2}$ do not share weights.

\subsection{Causal Global Prediction Model}

\begin{figure}[t]
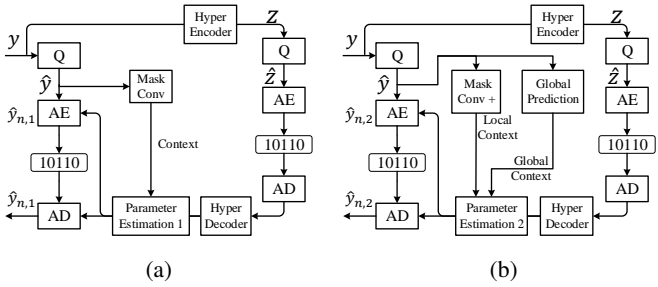

 \centering
 \begin{minipage}{0.46\columnwidth}
 \begin{subfigure}{\columnwidth}
 \includegraphics[scale=0.34, clip, trim=27.55cm 43.4cm 22cm 6.6cm]{figures/context_model.pdf}
 \caption{ \label{figure3a}}
 \end{subfigure}
 \end{minipage}
 \hspace{0.01\linewidth}
 \begin{minipage}{0.50\columnwidth}
 \begin{subfigure}{\columnwidth}
 \includegraphics[scale=0.34, clip, trim=42.3cm 43.4cm 7cm 6.6cm]{figures/context_model.pdf}
 \caption{ \label{figure3b}}
 \end{subfigure}
 \end{minipage}

 \caption{(a) Hyper model for the encoding and decoding of the first group $\bm{\hat{y}_{n,1}}$. (b) Incorporation of the adjacent and global context for the second group $\bm{\hat{y}_{n,2}}$.}
\label{figure3}
\end{figure}

The above causal context model is helpful for capturing both the spatial and channel redundancies. However, it only extracts local correlations but ignores global correlations. In this section, we improve it and propose a \textit{causal global prediction model} to utilize the long-range correlations of the latents. The proposed causal global prediction model separates the latents $\bm{\hat{y}}$ into two groups as well. The compression process of the first channel group $\bm{\hat{y}_{n,1}}$ is unchanged, only utilizing hyperprior and local context model, as illustrated in Fig.\ref{figure3a}. However, to estimate the distribution of the second latent group, the entropy model now incorporates both the improved adjacent context $\bm{c_{n,2}}$ and the global context $\bm{c_{n,3}}$ together as
\begin{equation}
\begin{aligned}
& \bm{c_{n,1}} = f_{c,1}(\bm{\hat{y}_{i_1}},\bm{\hat{y}_{i_2}},...,\bm{\hat{y}_{i_k}}), \\
& P_{\bm{\hat{y}_{n,1}|\hat{z}}}(\bm{\hat{y}_{n,1}|\bm{\hat{z}}})  \leftarrow h_{s,1}(\bm{\hat{z}},  \bm{c_{n,1}}| \bm{\theta_h}), \\
& \bm{c_{n,2}} = f_{c,2}(\bm{\hat{y}_{i_1}},\bm{\hat{y}_{i_2}},...,\bm{\hat{y}_{i_k}}, \bm{\hat{y}_{n,1}}), \\
& \bm{c_{n,3}} = f_{c,3}(\bm{\hat{y}_{1}},\bm{\hat{y}_{2}},...,\bm{\hat{y}_{n-1}}, \bm{\hat{y}_{n,1}}), \\
& P_{\bm{\hat{y}_{n,2}|\hat{z}}}(\bm{\hat{y}_{n,2}|\bm{\hat{z}}})  \leftarrow h_{s,2}(\bm{\hat{z}},  \bm{c_{n,2}}, \bm{c_{n,3}} | \bm{\theta_h}),
\end{aligned}
\label{equ7}
\end{equation}
where the improved adjacent context $\bm{c_{n,2}}$ is generated by the causal context model $f_{c,2}$ and global context $\bm{c_{n,3}}$ is generated by the causal global prediction model $f_{c,3}$. The causal global prediction model enables the decoder to establish accurate global reference information via latent separation, which will be introduced in the next section. Fig.\ref{figure3b} presents the diagram of the entropy coding process for the second group.

\begin{figure}[t]

 \centering
 \begin{subfigure}{\columnwidth}
 \includegraphics[scale=0.19, clip, trim=22.2cm 34.8cm 1.4cm 5.5cm]{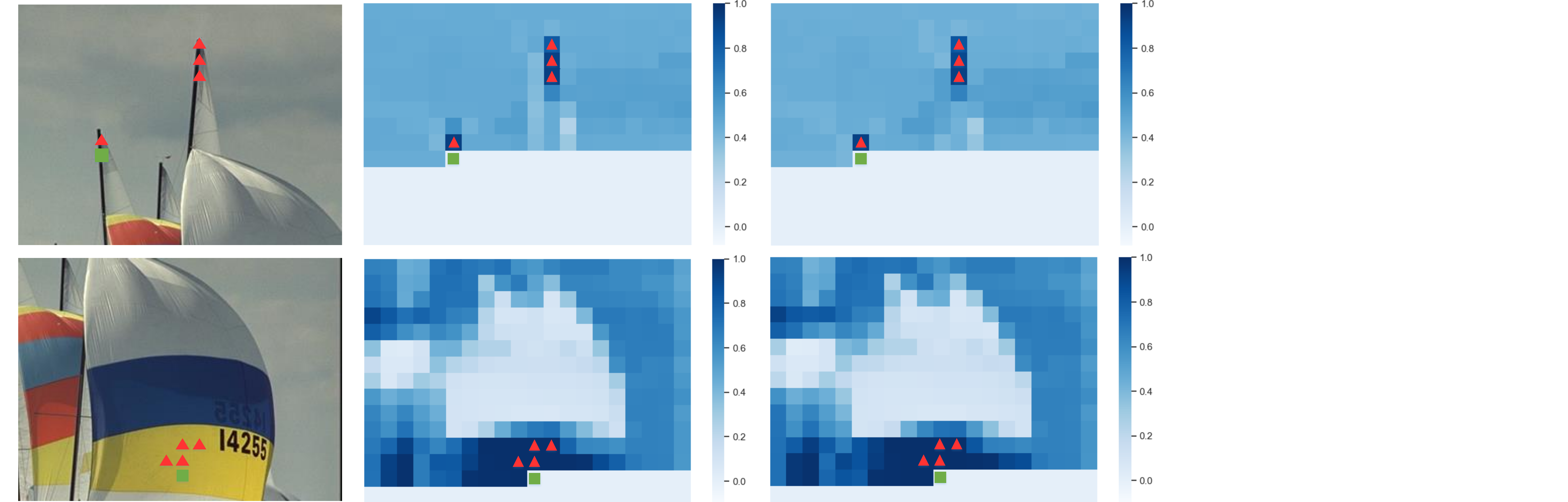}
\caption{Kodim09}
 \end{subfigure}
 \begin{subfigure}{\columnwidth}
 \includegraphics[scale=0.204, clip, trim=25.4cm 9cm 1.6cm 26cm]{figures/appendix_correlation_visual.pdf}
\caption{Kodim07}
 \end{subfigure}
\caption{From left to right: original image patch, correlation matrix of the entire latents, correlation matrix of the first half latents. It is observed that the global correlations of the entire latents can be approximated by the correlation matrices of the first half channels. \textcolor{green}{Green labels} represent the curent decoding point. \textcolor{red}{Red labels} denote to the four selected reference points.}
\label{figure4}
\end{figure}

\begin{figure*}[t]
 \centering
 \includegraphics[scale=0.81, clip, trim=2cm 9.8cm 7cm 3.2cm]{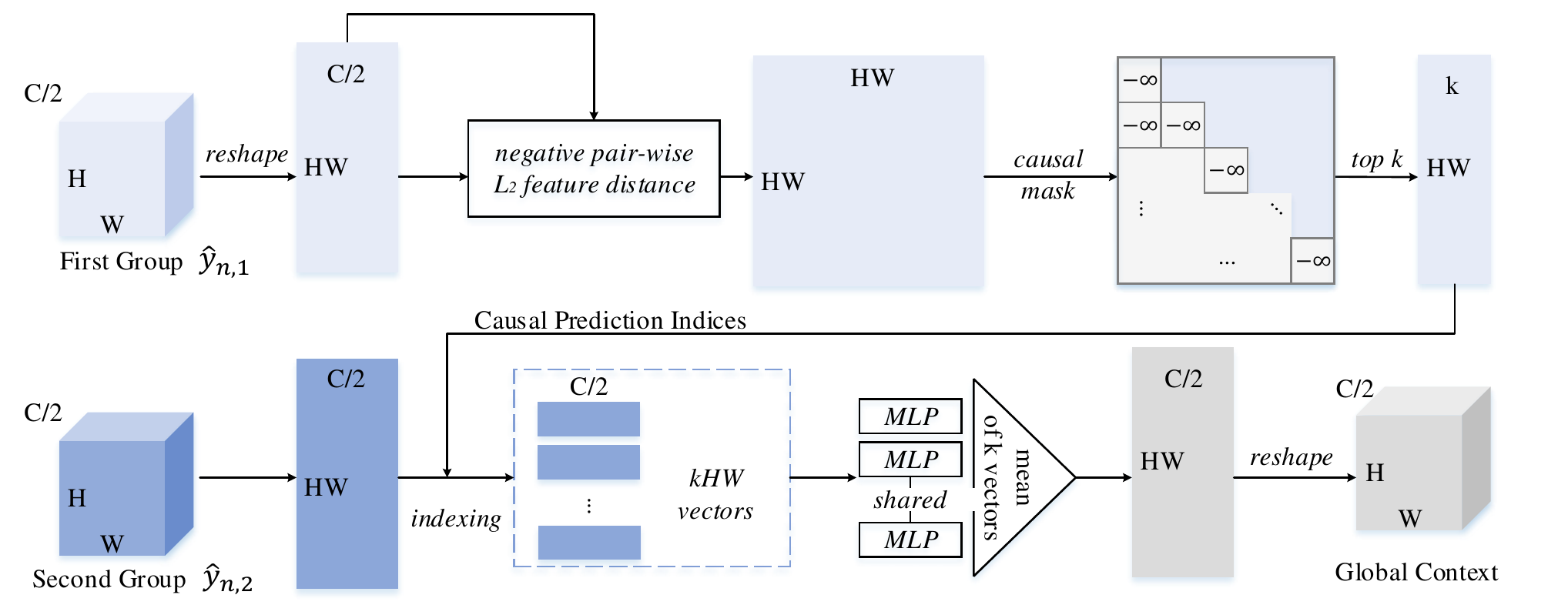}
 \caption{Global prediction via latent separation.}
\label{figure5}
\end{figure*}

\subsection{Global Prediction via Latent Separation}
The previous discussions in Section \ref{section3b} show the potential of global context for entropy estimation under the premise that the decoder can establish accurate global reference information. However, it is very bit-consuming to explicitly transmit the coordinates of relevant points.
As shown in Fig. \ref{figure4}, we find that the similarities among the half latent vectors match the similarities among the entire latent vectors. It implies that the spatial correlations of the entire latent variables can be approximated by the correlation matrices that are calculated from the first half channels. This observation motivates us to generate bit-free global correlations at the decoder side by making use of the concept of latent separation. The latent vector in each spatial location can roughly represent the context in a 16$\times$16 area of original image because the encoding transform network downsamples natural image for four times with stride=2. Therefore, there is a close relationship between the latent correlation matrix and the image context.

Fig.\ref{figure5} illustrates the process of separate global prediction. After decoding the first latent group \bm{$\hat{y}_{n,1}$}, we then calculate the negative pair-wise $\mathcal{L}_2$ distances among all half vectors, yielding a $HW \times HW$ correlation matrix. Since the decoding process is \textit{causal}, the correlation matrix is masked to be an upper triangular matrix. To avoid occasional inaccurate predictions, we preserve only the top-$k$ correlated points and gather their indices to generate the causal prediction indices. These indices, denoting the $k$ most similar points regarding the current decoding point, are used for indexing and selecting appropriate features in the second group. 
After indexing, we have $k$ vectors for each point, which will be utilized to predict current decoding point. The dimension of each prediction vector is $1\times 1 \times c/2$. Considering that the latents have a total of $HW$ points, we now have $kHW$ vectors as predicted information. Since current prediction vectors are directly gathered from previously decoded points, we then send these vectors into a shared multi-layer perceptron (MLP), which is a common process for learning effective representations \cite{liu2019learning}. We finally average the output of MLP to generate global context. In our experiments, we use top-4 predictions to generate global context which will be investigated in the following analysis. We also visualize some examples of top-4 predictions in Fig.\ref{figure4}.

\begin{figure}[t]
	\centering
 \begin{subfigure}{\columnwidth}
 \includegraphics[scale=0.26, clip, trim=1.4cm 0cm 1cm 1.4cm]{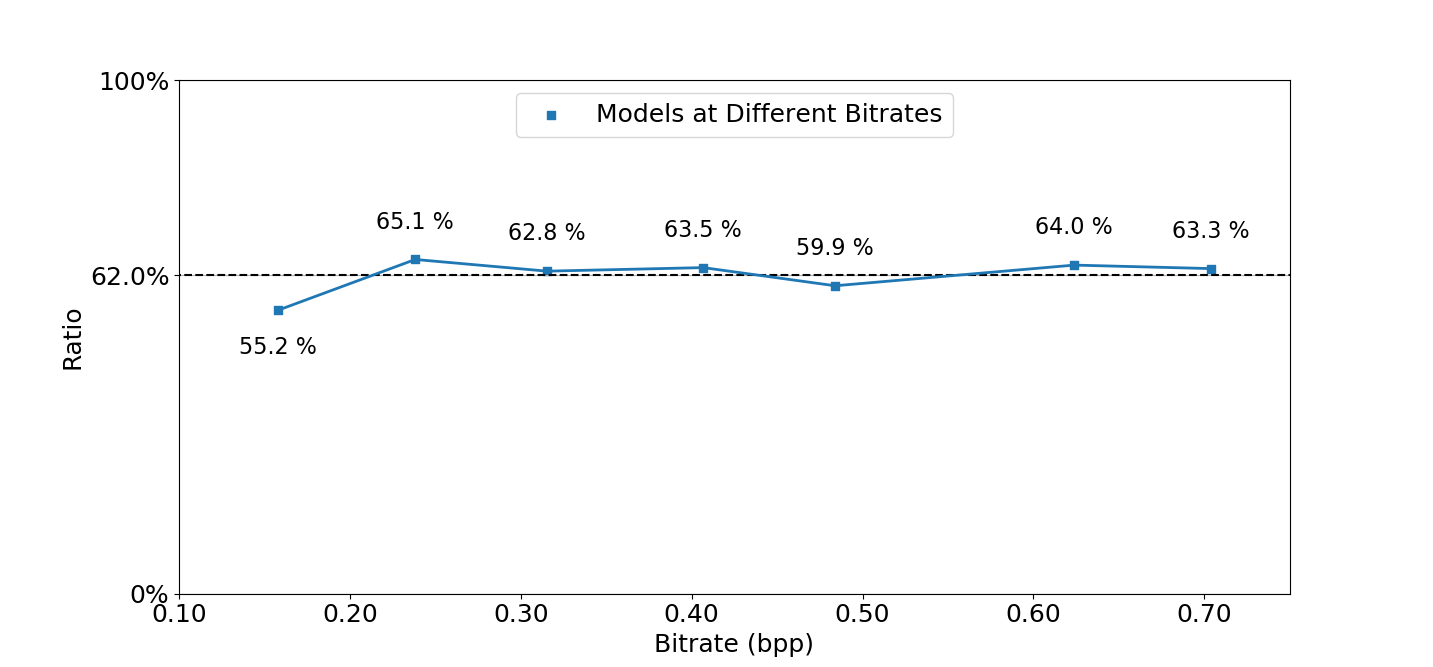}
 \caption{ \label{figure6a}}
 \end{subfigure}

 \begin{subfigure}{\columnwidth}
 \includegraphics[scale=0.26, clip, trim=1.4cm 0cm 1cm 1.4cm]{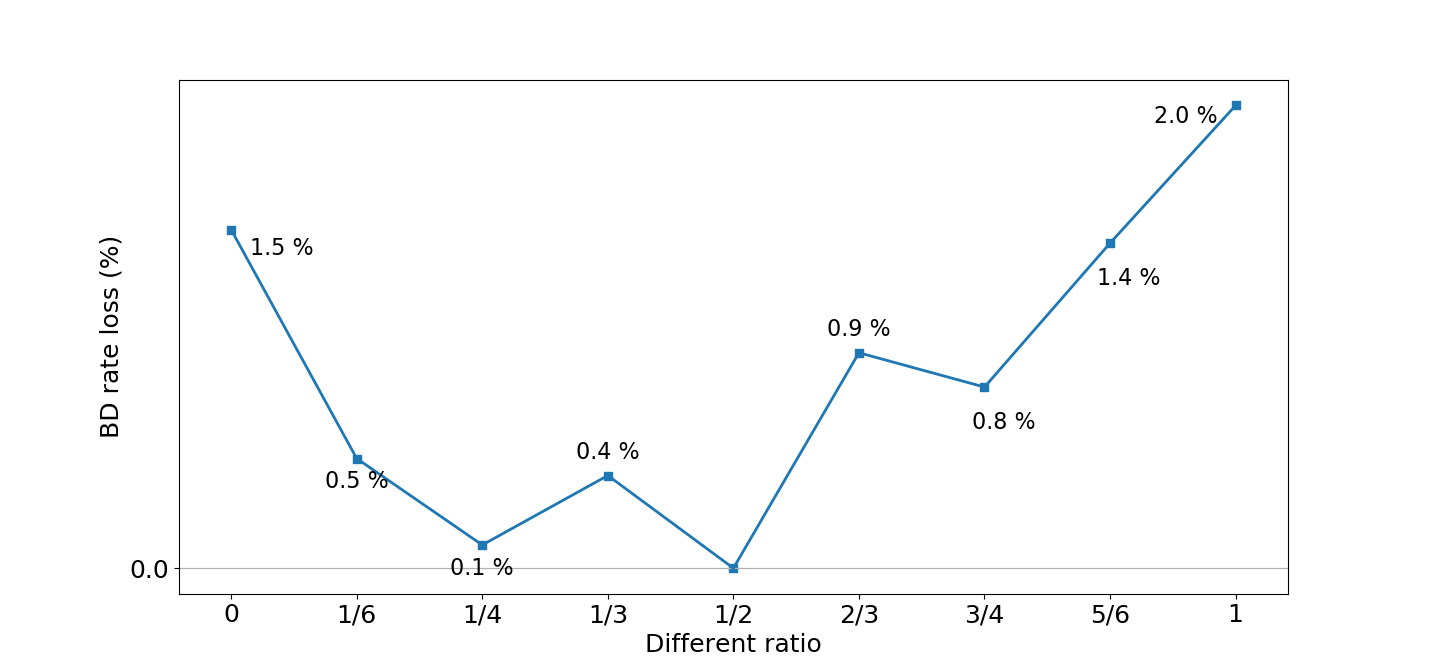}
 \caption{ \label{figure6b}}
 \end{subfigure}

\caption{(a) If separating the latents into two groups on average, the first group is representative that occupies a higher percentage in bitstream. (b) Different separation ratios affect the BD rate performance. We evaluate different methods on the whole Kodak dataset. The baseline here is the simplified model \cite{minnen2018joint} deployed with our proposed separate entropy coding modules.}
\label{figure6}
\end{figure}

\subsection{Analysis \label{section4d}}

\subsubsection{Why separating the latents into two groups on average}
The network is expected to adaptively arrange appropriate channels in different groups during training. As shown in Fig.\ref{figure6a}, when the first and second groups have the same number of channels, the first group accounts for around 62\% of the bitstream at different bitrates. This statistic demonstrates that to accurately model global correlations, the proposed latent-separated models are more likely to assign those representative and bit-consuming channels to the first group. We also explore to adjust the ratio of channel numbers in different groups. The results are shown in Fig.\ref{figure6b}. 
In this figure, the left-most point represents the case of a small ratio value where we assign two channels in the first group to calculate correlation matrix. These two channels in the first group are informative and representative of image contexts to guide the prediction of the second channel group. And ratio=1 is another extreme case that makes the whole context model degrade to conventional 2-D context model \cite{minnen2018joint} without channel separation. It is observed that the model achieves good results when the first channel group occupies 1/6$\sim$1/2 channels. Too small or too large ratio degrades compression performance since the concept of separate entropy coding requires some channels in the first group to ease the compression of the rest channels in the second group. In particular, ratio$\approx$0 performs better than ratio=1 slightly since the former can still utilize coarse correlation information to achieve more effective compression.

In addition, it is reasonable to simply separate the latent channels into two groups. On the one hand, the parameter estimation modules in different groups do not share weights. On the other hand, the current two-group separation already divides the decoding period into two stages and thus increases time complexity. Dividing the latent channels into more groups would further complicate the decoding in terms of both space complexity and time complexity. 

\subsubsection{Difference between global context and local context}
The global context should be distinguished from the local context. Local context \cite{minnen2018joint,lee2019context} is extracted by a mask convolution layer, \textit{aggregating} possible context without performing explicit prediction. But here the causal global prediction model searches the set of all decoded points, selecting the top-k correlated points. These selected points will then take their second channel group as references to \textit{predict} the undecoded channel group in current decoding point.

\subsubsection{Difference between global context and hyperprior model}
The generated global context is also different from hierarchical priors. Both of them take advantage of the global structures of the latents to improve entropy estimation. However, the hyperprior model transmits additional overhead, i.e., side information, to model rough global structures with downsampling and upsampling layers. Our causal global prediction model does not transmit overhead but still enables effective global prediction with all accessible points.
In terms of compression performance, global context is complementary to hierarchical priors, even if they only work for the second channel group. However, similar to local context model \cite{minnen2018joint}, latent-separated global prediction results in a causality problem for decoding that increases decoding complexity.

\begin{figure}[t]
	\centering
     \includegraphics[scale=0.39, clip, trim=4.3cm 2.4cm 2.6cm 5cm]{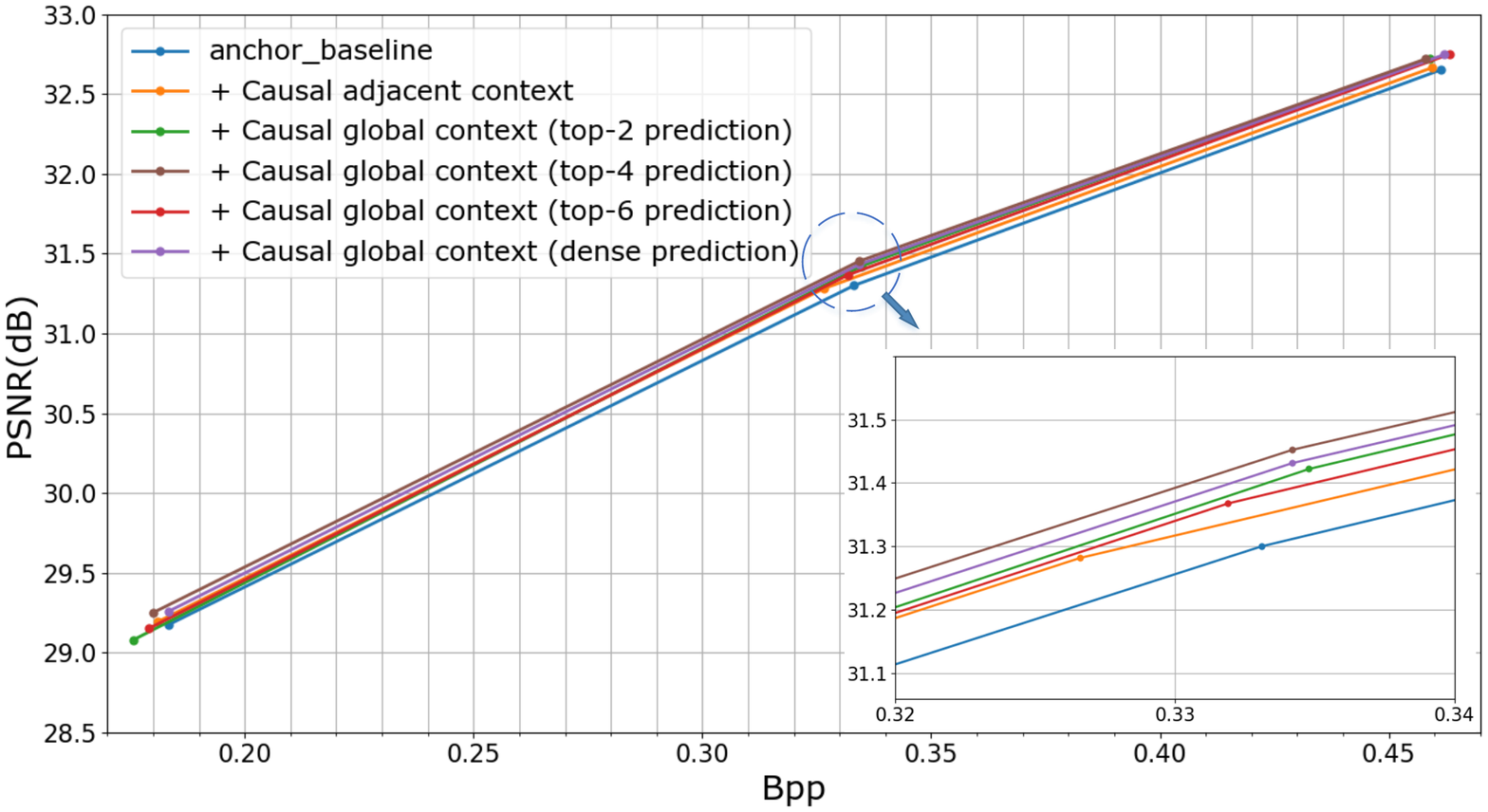}
\caption{Exploring the effects of different $k$ values in our proposed causal global prediction model.}
\label{figure7}
\end{figure}

\subsubsection{The influence of the value of k}
The proposed causal global prediction model selects the $k$ most relevant points as references to predict the entropy of undecoded channel group. It is important to determine an appropriate number of reference points because it directly influences the prediction performance. Actually, in traditional compression standards such as H.264 \cite{wiegand2003overview}, different prediction modes also involve different numbers of reference points. For example, vertical prediction mode and horizontal prediction mode directly copy one adjacent pixel. Some diagonal prediction modes would involve two relevant adjacent pixels. For DC prediction mode and planar mode \cite{wiegand2003draft}, they operate on more adjacent pixels. Here, we explore the effects of the number of reference points on our learned image compression method.

As shown in Fig.\ref{figure7}, we compare different $k$ values, where baseline is \cite{minnen2018joint}. In this figure, dense prediction refers to using all decoded points to predict the current point while top-$2,4,6$ prediction only involves using several relevant points for prediction. It is found that $k=4$ achieves the best performance, which is slightly better than dense prediction. This can be explained by the fact that some global references are inaccurate because the global correlation matrix is approximated by half channels. We emphasize that the separate global prediction model is not a simple non-local attention \cite{wang2018non}. In the task of compression, the decoder cannot explicitly calculate attention map at the decoder side. Furthermore, the proposed causal global prediction model is also different from the causal non-local attention \cite{chen2018pixelsnail} used for image distribution modeling. Our method determines accurate reference points, while the causal non-local attention simply aggregates all previous points and cannot establish spatial correlations.

\subsubsection{Analysis of cross-channel redundancy}
We should note that the cross-channel redundancy still exists after non-linear transform and it is utilized with spatial correlation together. On the one hand, the transform network is not efficient enough to remove the cross-channel redundancy. In the non-linear transform network, one critical component is Generalized Divisive Normalization (GDN) layer that helps to decorrelate information across channels \cite{balle2017end}. This GDN layer achieves a small mutual information between transformed components as illustrated in \cite{balle2016density}. However, as shown in the figure 1 of \cite{balle2016density}, the mutual information after GDN transform is not completely zero, which implies that residual correlation exists after such non-linear transform. On the other hand, the cross-channel relationship is utilized together with spatial correlation. Assume that we have two same latent vectors in different spatial location. Now we want to use the first latent vector to predict the other undecoded latent vector. Even if the cross-channel redundancy was removed entirely, we can still determine the similarity between these two latent vectors by examining their first half vectors because their first half channels are exactly the same. We thus can conduct causal global prediction for the second channel group.

\subsection{Group-Separated Attention Module in Transforms}
\begin{figure}[t]
	\centering
 	\begin{subfigure}{\columnwidth}
     \includegraphics[scale=0.62, clip, trim=28.8cm 23.8cm 14cm 14.3cm]{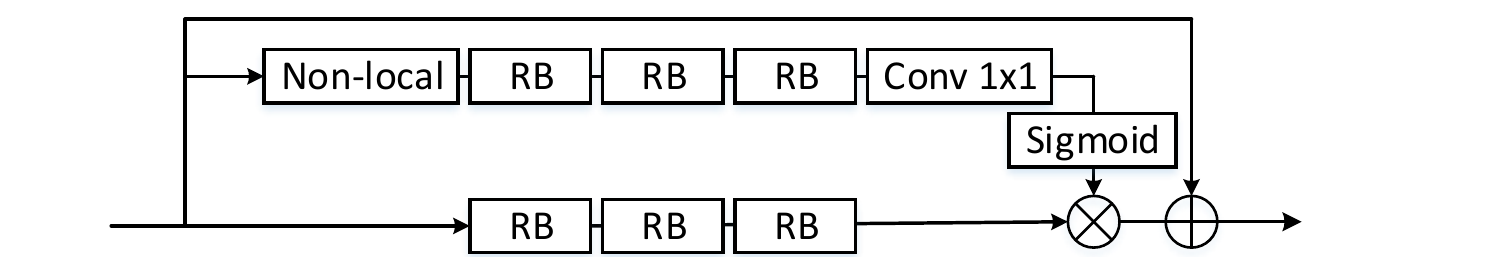}
 	\caption{Non-local version in \cite{chen2019neural}. \label{figure8a}}
 	\end{subfigure}
 	\begin{subfigure}{\columnwidth}
     \includegraphics[scale=0.62, clip, trim=28.8cm 20cm 14cm 17.8cm]{figures/frequency_split.pdf}
 	\caption{Simplified version in \cite{cheng2020learned}. \label{figure8b}}
 	\end{subfigure}
 	\begin{subfigure}{\columnwidth}
     \includegraphics[scale=0.48, clip, trim=28.0cm 12.2cm 13cm 21.1cm]{figures/frequency_split.pdf}
 	\caption{Our proposed group-separated attention. \label{figure8c}}
 	\end{subfigure}
\caption{Comparisons of different versions of attention. RB represents residual block, which contains three $3\times 3$ convolution layers (stride$=$1) and three ReLU activation layers.}
\label{figure8}
\end{figure}

\begin{figure*}[h]
 \centering
 \includegraphics[scale=0.35, clip, trim=6cm 15.7cm 1.4cm 24.4cm]{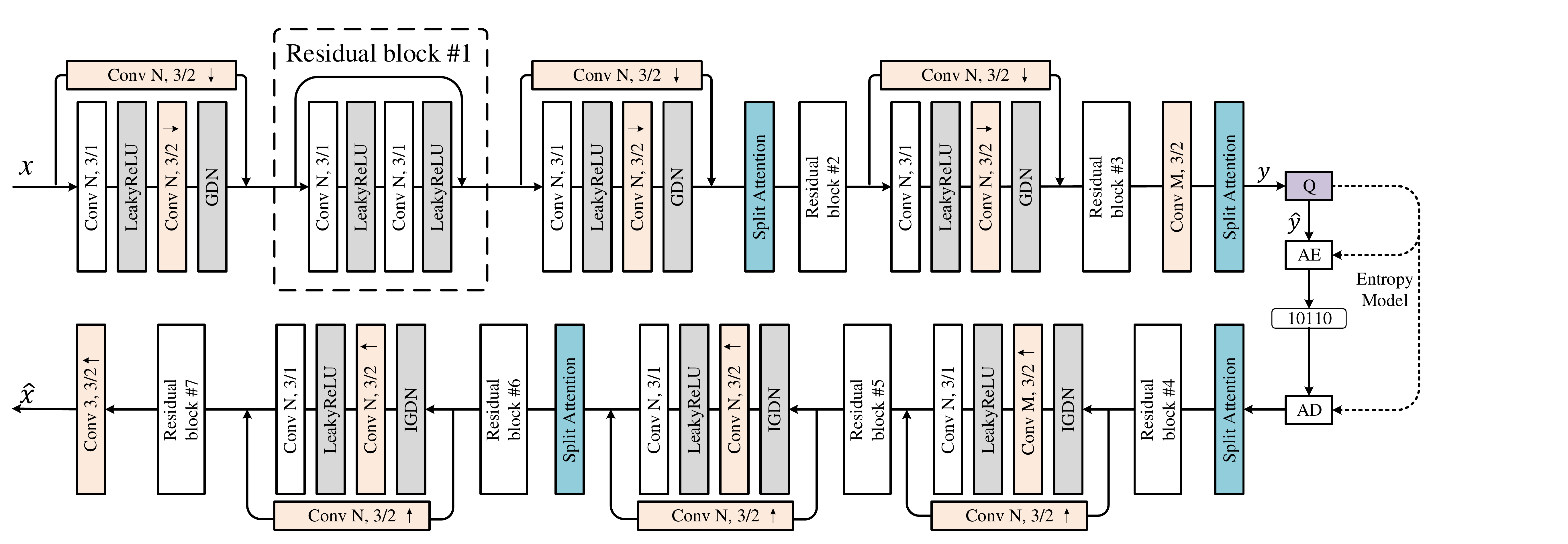}
 \caption{Architecture of our main compression network. We choose N=192, M=192 when bitrates are less than 0.50 bpp. For higher bitrate, we set N=192, M=320 to enhance the model capacity following the setting in \cite{balle2018variational}.}
\label{figure9}
\end{figure*}
In the previous sections, to reduce the cross-channel redundancies and global redundancies among the latents, we introduce the novel causal context model and causal global prediction model. Both of them facilitate entropy estimation for the second latent group. 
In addition to the improved entropy model, we further propose a new group-separated attention layer to enhance the non-linear transform networks.


There are several previous works with respect to learned image compression using attention modules to enhance the encoder and decoder. Chen et al. \cite{chen2019neural} suggests employing a residual non-local module for compression. Later, Cheng et al. simplify this attention layer \cite{cheng2020learned} by removing the non-local block with comparable performance. In this paper, inspired by \cite{zhang2020resnest,zhang2020multi}, we empirically modify the attention structure and adopt an improved group-separated attention module, which enables separate feature-map attention in two groups. 
The channel splitting process is expected to gather those channels with similar characteristics.
Detailed comparisons between these three types of attention can be found in Fig.\ref{figure8}. Experimental results indicate that this group-separated attention module is more powerful than other attention mechanisms (shown later in the experimental section).

\subsection{Network Architecture}

The architecture of our main compression network is similar as \cite{cheng2020learned}. Our main contributions are to employ the novel causal context model and causal global prediction model and replace the original attention module with a separate attention module. Besides, we enhance the parameter estimation module using residual blocks \cite{he2016deep} as described in our pioneer work \cite{guo20203}. The main network structure is shown in Fig.\ref{figure9}. To build a powerful compression model, we integrate a postprocessing network GRDN \cite{kim2019grdn} as suggested in \cite{lee2019hybrid}. This postprocessing network is lightweight and achieves a good balance between complexity and performance.
To estimate the entropy of $\hat{y}$, we assume the distribution of $y$ subject to Gaussian mixture model (GMM) and this is a generalized form to approximate arbitrary probability distribution. Mathematically, 
\begin{equation}
\begin{aligned}
p(\bm{\hat{y}}|\bm{\hat{z}}) & \sim \sum_{i=1}^{I} \pi_{i}  \mathcal N(\mu_{i}, \sigma^2_{i}), \\
\pi_{i},\mu_{i}, \sigma^2_{i} & \leftarrow h_s(\bm{\hat{z}},  (\bm{c_{n,1}}, \bm{c_{n,2}}, \bm{c_{n,3}}) | \bm{\theta_h})
\end{aligned}
\label{equ8}
\end{equation}
where $\pi_{i},\mu_{i}, \sigma_{i},i\in\{1,...,I\}$ are the estimated parameters of the entropy model (we choose $I=3$ following \cite{cheng2020learned,guo20203}). Eq.\ref{equ8} models a continuous probability density regarding $y$. To estimate the discrete probability of $\hat{y}$, the continuous function $p(y|\hat{z})$ is convolved with a unit uniform distribution, which is usually implemented in the form of additive uniform noise,
\begin{equation}
\begin{aligned}
P_{\hat{y}|\hat{z}}(\bm{\hat{y}|\bm{\hat{z}}}) & =  (\sum_{i=1}^{I} \pi_{i}  \mathcal N(\mu_{i}, \sigma^2_{i}) \ast \mathcal{U}(-\frac{1}{2}, \frac{1}{2}))(\bm{\hat{y}}).
\end{aligned}
\label{equ9}
\end{equation}
As explained in the work of \cite{balle2017end}, such additive uniform noise helps to solve the problem of non-differential quantization. Note that we sent the hard-quantized latent variable $\hat{y}$ into the synthesis decoder during training, which is similar to \cite{minnen2020channel} and is found to achieve better performance. We describe the experimental results in the following section.

\section{Experiments\label{section5}}

\subsection{Implementation Details}

\begin{figure}[t]

    \begin{subfigure}{\columnwidth}
	\centering
     \includegraphics[scale=0.31, clip, trim=1.5cm 0.3cm 2cm 2.0cm]{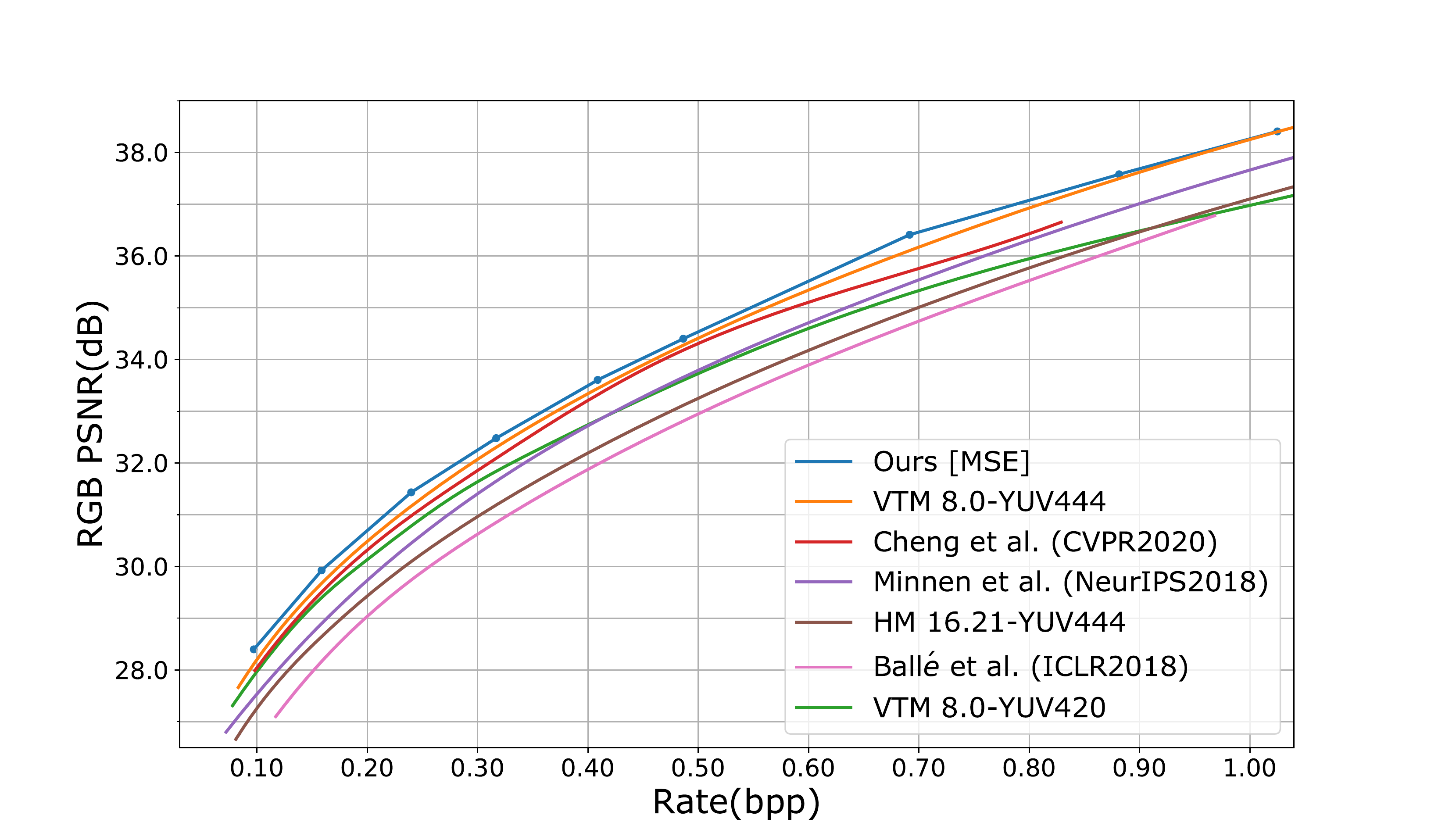}
    \end{subfigure}

    \begin{subfigure}{\columnwidth}
	\centering
     \includegraphics[scale=0.31, clip, trim=1.5cm 0.3cm 2cm 2.0cm]{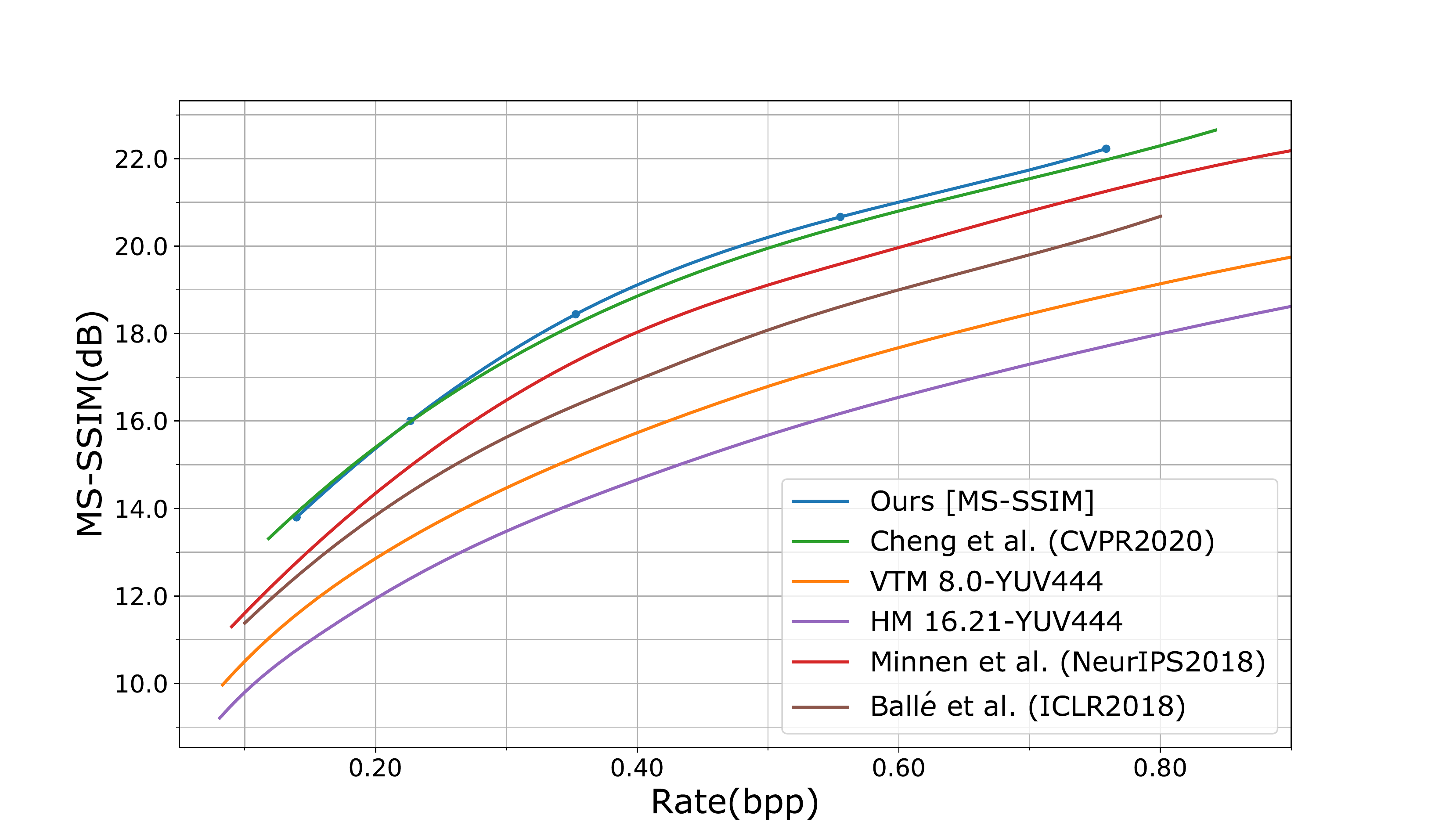}
    \end{subfigure}

\caption{Comparisons of RD curves with other methods.}
\label{figure10}
\end{figure}

Our compression model is trained on the whole ImageNet training set \cite{deng2009imagenet}. The original images are cropped to $256 \times 256$ patches. Minibatches of 8 of these patches are used to update network parameters. We divide the training period into three stages. First, we train the main compression network with rate-distortion constraint (as in Eq.\ref{equ2}). This training stage lasts for 1.5 million iterations. Second, we fix the main compression network and optimize the postprocessing GRDN to minmize distortion, which lasts for 400,000 iterations. Third, both the main network and the postprocessing network are optimized jointly to achieve the optimal rate-distortion performance. The third training stage lasts for 600,000 iterations.
For all three training stages, we use the Adam optimizer \cite{kingma2014adam} with an initial learning rate of $5e-5$. The learning rate decays to $1e-5$ after 300,000 iterations. 

Similar to previous works, the distortion term is measured by two quality metrics, PSNR and MS-SSIM \cite{wang2004image}. 
During training, a Lagrange multiplier $\lambda$ balances the rate-distortion trade-off. When optimizing for PSNR, we train different models with values of $\lambda$ ranging from 100 to 3072, where $D=MSE(\hat{x}, x)$. In terms of MS-SSIM, we train five models with $\lambda \in \{6, 16, 40, 100, 180\}$, where $D=1-$MSSSIM$(\hat{x}, x)$. 

\subsection{Performance}

For comparison, we evaluate different methods on the Kodak dataset which contains 24 uncompressed images. The rate distortion curves are shown in Fig.\ref{figure10}. We compare the performance of our method with other existing approaches including \cite{cheng2020learned}, \cite{minnen2018joint}, \cite{balle2018variational} and traditional codecs, such as VVC intra and HEVC intra. Specifically, the results of Ball\'e et al. (ICLR2018) \cite{balle2018variational} are obtained from our reproduced model, which performs closely as their report. The results of Minnen et al. (NeurIPS2018) \cite{minnen2018joint} and Cheng et al. (CVPR2020) \cite{cheng2020learned} are directly taken from their papers because our reproduced statistics are slightly lower than their report. The reproduction gap may be caused by differences in the training set. While we directly use the whole ImageNet training set, \cite{minnen2018joint} does not mention the training set and \cite{cheng2020learned} uses a subset of ImageNet to avoid negative samples.
For VVC and HEVC, we use the official test model VTM 8.0 and HM 16.21\footnote{\url{https://vcgit.hhi.fraunhofer.de/jct-vc/HM/-/releases/HM-16.21}}, with the YUV444 configuration in all-intra mode. 

\begin{figure*}[t]
\centering
    \begin{minipage}{0.236\textwidth}
    \begin{subfigure}{\columnwidth}
     \includegraphics[scale=0.153]{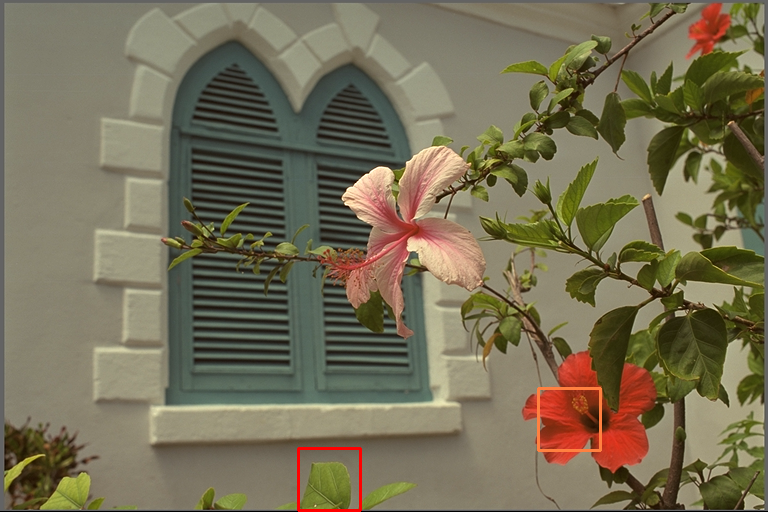}
    \end{subfigure}
    \end{minipage}
    \begin{minipage}{0.07\textwidth}
    \begin{subfigure}{\columnwidth}
     \includegraphics[scale=0.60]{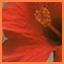}
    \end{subfigure}
\\[1pt]
    \begin{subfigure}{\columnwidth}
     \includegraphics[scale=0.60]{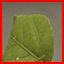}
    \end{subfigure}
    \end{minipage}
 \hspace{0.008\linewidth}
    \begin{minipage}{0.236\textwidth}
    \begin{subfigure}{\columnwidth}
     \includegraphics[scale=0.153]{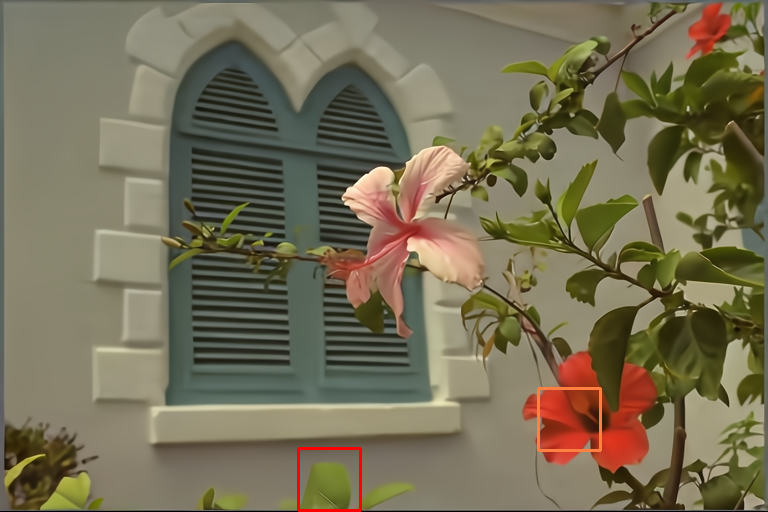}
    \end{subfigure}
    \end{minipage}
    \begin{minipage}{0.07\textwidth}
    \begin{subfigure}{\columnwidth}
     \includegraphics[scale=0.60]{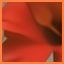}
    \end{subfigure}
\\[1pt]
    \begin{subfigure}{\columnwidth}
     \includegraphics[scale=0.60]{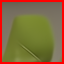}
    \end{subfigure}
    \end{minipage}
 \hspace{0.008\linewidth}
    \begin{minipage}{0.236\textwidth}
    \begin{subfigure}{\columnwidth}
     \includegraphics[scale=0.153]{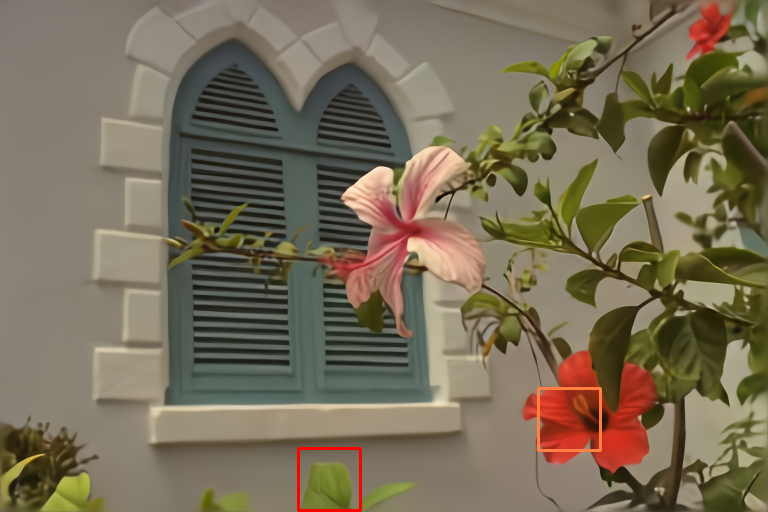}
    \end{subfigure}
    \end{minipage}
    \begin{minipage}{0.07\textwidth}
    \begin{subfigure}{\columnwidth}
     \includegraphics[scale=0.60]{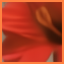}
    \end{subfigure}
\\[1pt]
    \begin{subfigure}{\columnwidth}
     \includegraphics[scale=0.60]{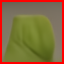}
    \end{subfigure}
    \end{minipage}
\\[2pt]
\begin{minipage}{0.306\textwidth}
    \begin{subfigure}{\columnwidth}
	\caption*{Ground Truth}
    \end{subfigure}
\end{minipage}
 \hspace{0.008\linewidth}
\begin{minipage}{0.306\textwidth}
    \begin{subfigure}{\columnwidth}
	\caption*{Ours [MSE] \\ 0.136 bpp, 32.36dB, 0.9761}
    \end{subfigure}
\end{minipage}
\hspace{0.008\linewidth}
\begin{minipage}{0.33\textwidth}
    \begin{subfigure}{\columnwidth}
	\caption*{Ours [MS-SSIM] \\  0.105 bpp, 27.66dB, 0.9750}
    \end{subfigure}
\end{minipage}


    \begin{minipage}{0.236\textwidth}
    \begin{subfigure}{\columnwidth}
     \includegraphics[scale=0.153]{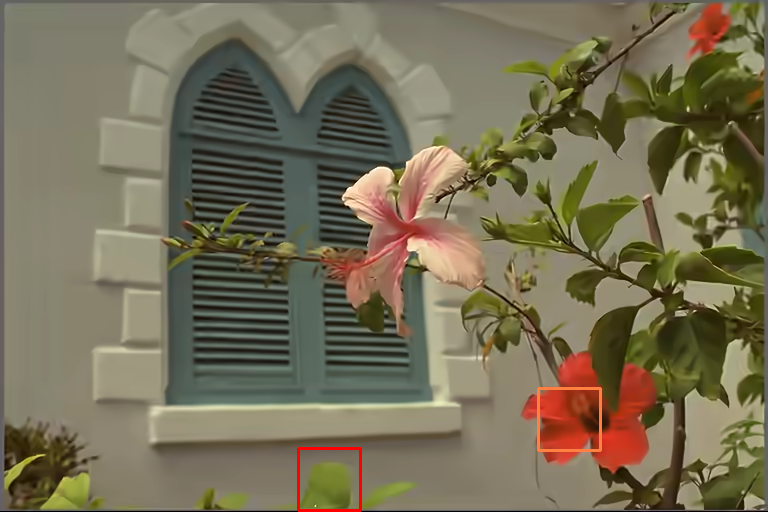}
    \end{subfigure}
    \end{minipage}
    \begin{minipage}{0.07\textwidth}
    \begin{subfigure}{\columnwidth}
     \includegraphics[scale=0.60]{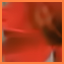}
    \end{subfigure}
\\[1pt]
    \begin{subfigure}{\columnwidth}
     \includegraphics[scale=0.60]{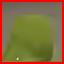}
    \end{subfigure}
    \end{minipage}
 \hspace{0.008\linewidth}
    \begin{minipage}{0.236\textwidth}
    \begin{subfigure}{\columnwidth}
     \includegraphics[scale=0.153]{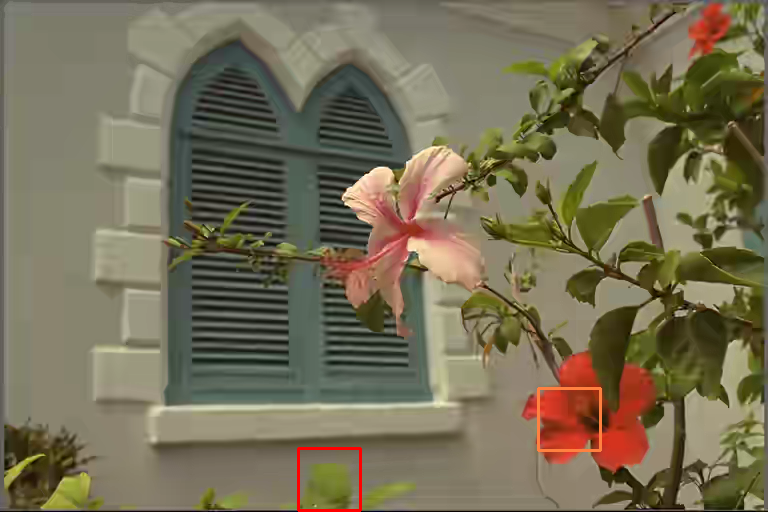}
    \end{subfigure}
    \end{minipage}
    \begin{minipage}{0.07\textwidth}
    \begin{subfigure}{\columnwidth}
     \includegraphics[scale=0.60]{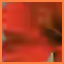}
    \end{subfigure}
\\[1pt]
    \begin{subfigure}{\columnwidth}
     \includegraphics[scale=0.60]{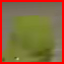}
    \end{subfigure}
    \end{minipage}
 \hspace{0.008\linewidth}
    \begin{minipage}{0.236\textwidth}
    \begin{subfigure}{\columnwidth}
     \includegraphics[scale=0.153]{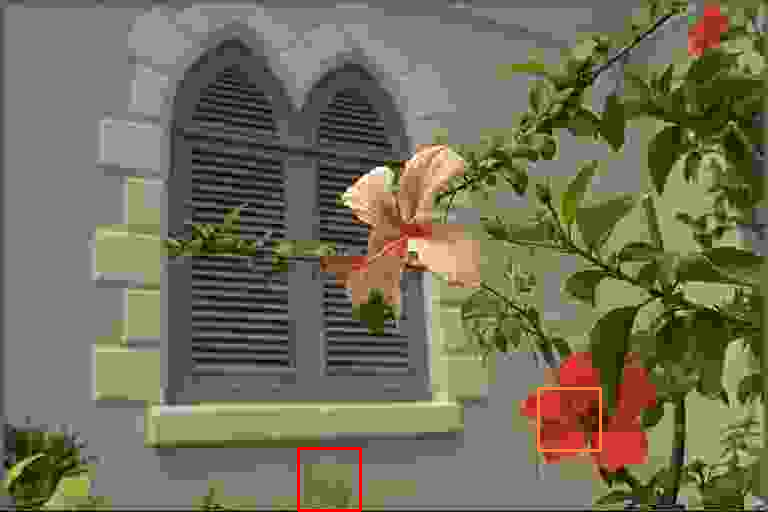}
    \end{subfigure}
    \end{minipage}
    \begin{minipage}{0.07\textwidth}
    \begin{subfigure}{\columnwidth}
     \includegraphics[scale=0.60]{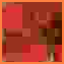}
    \end{subfigure}
\\[1pt]
    \begin{subfigure}{\columnwidth}
     \includegraphics[scale=0.60]{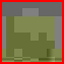}
    \end{subfigure}
    \end{minipage}
\\[2pt]
\begin{minipage}{0.306\textwidth}
    \begin{subfigure}{\columnwidth}
	\caption*{VTM 8.0 \\ 0.141 bpp, 31.63dB, 0.9702}
    \end{subfigure}
\end{minipage}
 \hspace{0.008\linewidth}
\begin{minipage}{0.306\textwidth}
    \begin{subfigure}{\columnwidth}
	\caption*{HM 16.21 \\ 0.135 bpp, 29.94dB, 0.9573}
    \end{subfigure}
\end{minipage}
\hspace{0.008\linewidth}
\begin{minipage}{0.33\textwidth}
    \begin{subfigure}{\columnwidth}
	\caption*{JPEG \\ 0.199 bpp, 23.25dB, 0.8381}
    \end{subfigure}
\end{minipage}
\\[8pt]
   \begin{minipage}{0.236\textwidth}
    \begin{subfigure}{\columnwidth}
     \includegraphics[scale=0.153]{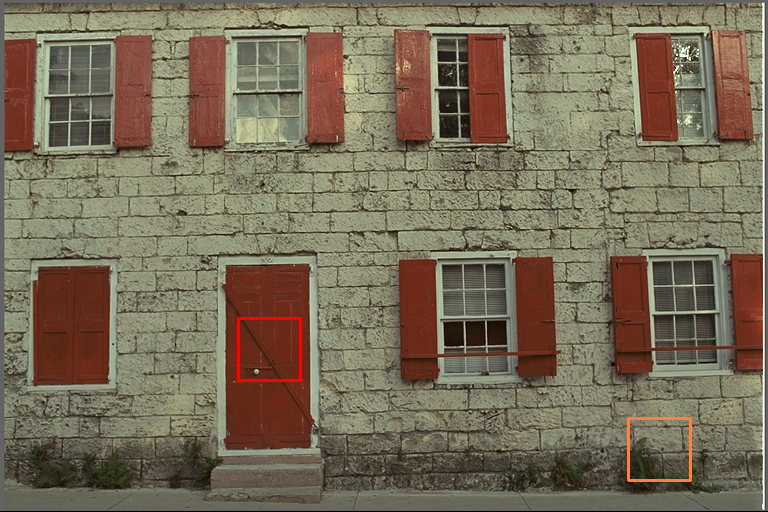}
    \end{subfigure}
    \end{minipage}
    \begin{minipage}{0.07\textwidth}
    \begin{subfigure}{\columnwidth}
     \includegraphics[scale=0.60]{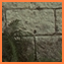}
    \end{subfigure}
\\[1pt]
    \begin{subfigure}{\columnwidth}
     \includegraphics[scale=0.60]{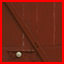}
    \end{subfigure}
    \end{minipage}
 \hspace{0.008\linewidth}
    \begin{minipage}{0.236\textwidth}
    \begin{subfigure}{\columnwidth}
     \includegraphics[scale=0.153]{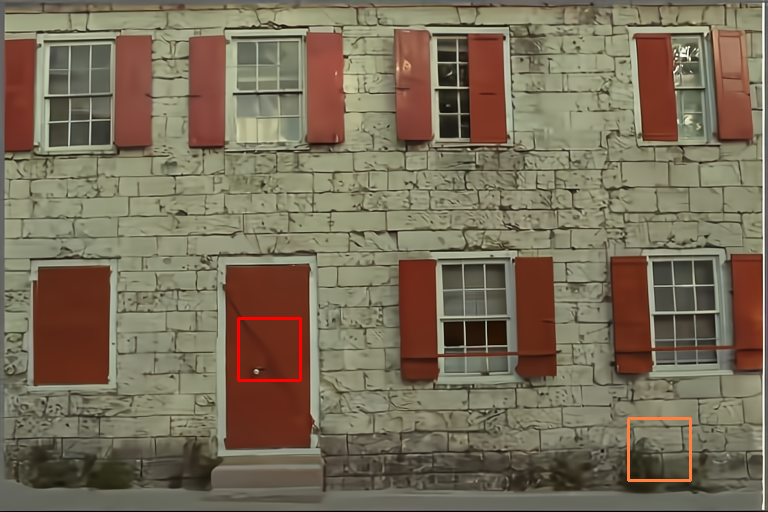}
    \end{subfigure}
    \end{minipage}
    \begin{minipage}{0.07\textwidth}
    \begin{subfigure}{\columnwidth}
     \includegraphics[scale=0.60]{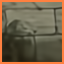}
    \end{subfigure}
\\[1pt]
    \begin{subfigure}{\columnwidth}
     \includegraphics[scale=0.60]{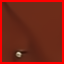}
    \end{subfigure}
    \end{minipage}
 \hspace{0.008\linewidth}
    \begin{minipage}{0.236\textwidth}
    \begin{subfigure}{\columnwidth}
     \includegraphics[scale=0.153]{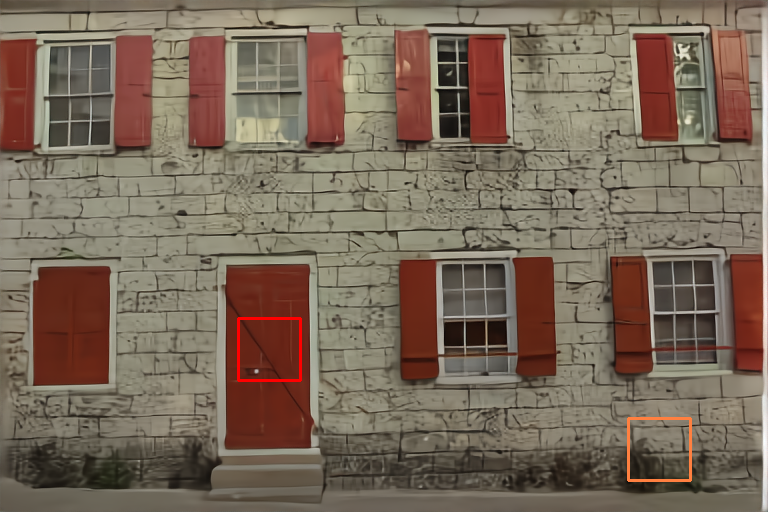}
    \end{subfigure}
    \end{minipage}
    \begin{minipage}{0.07\textwidth}
    \begin{subfigure}{\columnwidth}
     \includegraphics[scale=0.60]{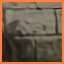}
    \end{subfigure}
\\[1pt]
    \begin{subfigure}{\columnwidth}
     \includegraphics[scale=0.60]{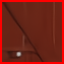}
    \end{subfigure}
    \end{minipage}
\\[2pt]
\begin{minipage}{0.306\textwidth}
    \begin{subfigure}{\columnwidth}
	\caption*{Ground Truth}
    \end{subfigure}
\end{minipage}
 \hspace{0.008\linewidth}
\begin{minipage}{0.306\textwidth}
    \begin{subfigure}{\columnwidth}
	\caption*{Ours [MSE] \\ 0.229 bpp, 27.05dB, 0.9428}
    \end{subfigure}
\end{minipage}
\hspace{0.008\linewidth}
\begin{minipage}{0.33\textwidth}
    \begin{subfigure}{\columnwidth}
	\caption*{Ours [MS-SSIM] \\  0.171 bpp, 23.28dB, 0.9436}
    \end{subfigure}
\end{minipage}


    \begin{minipage}{0.236\textwidth}
    \begin{subfigure}{\columnwidth}
     \includegraphics[scale=0.153]{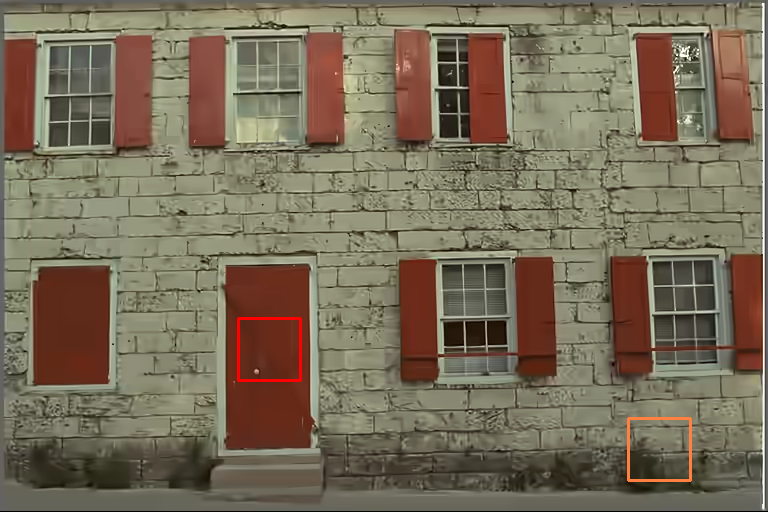}
    \end{subfigure}
    \end{minipage}
    \begin{minipage}{0.07\textwidth}
    \begin{subfigure}{\columnwidth}
     \includegraphics[scale=0.60]{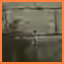}
    \end{subfigure}
\\[1pt]
    \begin{subfigure}{\columnwidth}
     \includegraphics[scale=0.60]{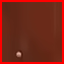}
    \end{subfigure}
    \end{minipage}
 \hspace{0.008\linewidth}
    \begin{minipage}{0.236\textwidth}
    \begin{subfigure}{\columnwidth}
     \includegraphics[scale=0.153]{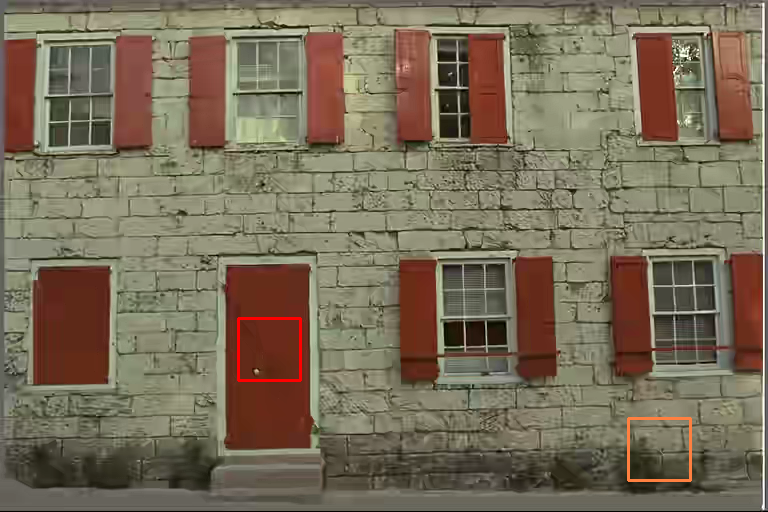}
    \end{subfigure}
    \end{minipage}
    \begin{minipage}{0.07\textwidth}
    \begin{subfigure}{\columnwidth}
     \includegraphics[scale=0.60]{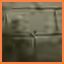}
    \end{subfigure}
\\[1pt]
    \begin{subfigure}{\columnwidth}
     \includegraphics[scale=0.60]{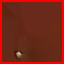}
    \end{subfigure}
    \end{minipage}
 \hspace{0.008\linewidth}
    \begin{minipage}{0.236\textwidth}
    \begin{subfigure}{\columnwidth}
     \includegraphics[scale=0.153]{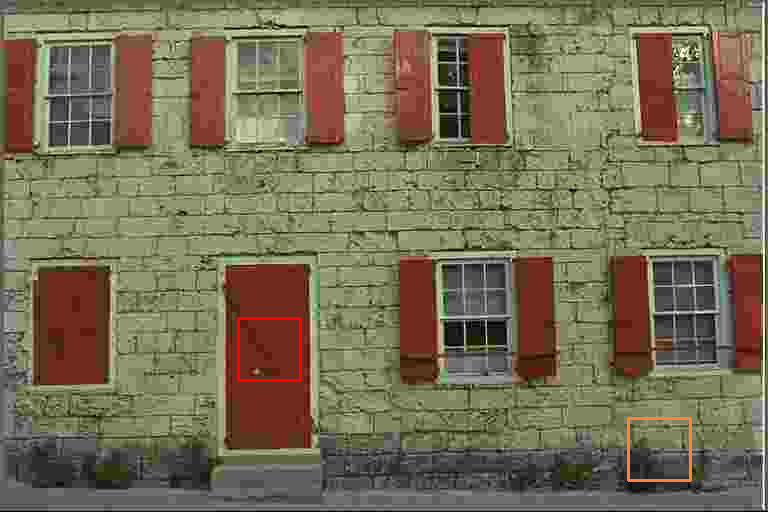}
    \end{subfigure}
    \end{minipage}
    \begin{minipage}{0.07\textwidth}
    \begin{subfigure}{\columnwidth}
     \includegraphics[scale=0.60]{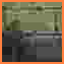}
    \end{subfigure}
\\[1pt]
    \begin{subfigure}{\columnwidth}
     \includegraphics[scale=0.60]{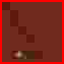}
    \end{subfigure}
    \end{minipage}
\\[2pt]
\begin{minipage}{0.306\textwidth}
    \begin{subfigure}{\columnwidth}
	\caption*{VTM 8.0 \\ 0.236 bpp, 26.94dB, 0.9377}
    \end{subfigure}
\end{minipage}
 \hspace{0.008\linewidth}
\begin{minipage}{0.306\textwidth}
    \begin{subfigure}{\columnwidth}
	\caption*{HM 16.21 \\ 0.238 bpp, 26.10dB, 0.9253}
    \end{subfigure}
\end{minipage}
\hspace{0.008\linewidth}
\begin{minipage}{0.33\textwidth}
    \begin{subfigure}{\columnwidth}
	\caption*{JPEG \\ 0.305 bpp, 22.89dB, 0.8568}
    \end{subfigure}
\end{minipage}
\caption{Visual comparisons.}
\label{figure11}
\end{figure*}

Fig.\ref{figure10} quantitatively shows that our method achieves the state-of-the-art rate-distortion performance. Specifically, it outperforms VTM 8.0 at all bitrates in terms of PSNR and performs better than the previous state-of-the-art method \cite{cheng2020learned} at most bitrates in terms of MS-SSIM. Our method achieves 5.1\% BD rate savings against VTM 8.0 (covering 0.15, 0.4, 0.7 and 1.0 bpp). There is an irregular point in our PSNR-rate curve (at 0.68 bpp), which is the turning point that indicates the change of model capacity. 
In Fig.\ref{figure11}, we provide some examples that exhibit pleasant qualitative results obtained by our method, especially the model optimized for MS-SSIM.

\begin{figure*}[t]
\centering
    \begin{minipage}{0.48\textwidth}
    \begin{subfigure}{\columnwidth}
\includegraphics[scale=0.31, clip, trim=1.5cm 0.3cm 2cm 2.0cm]{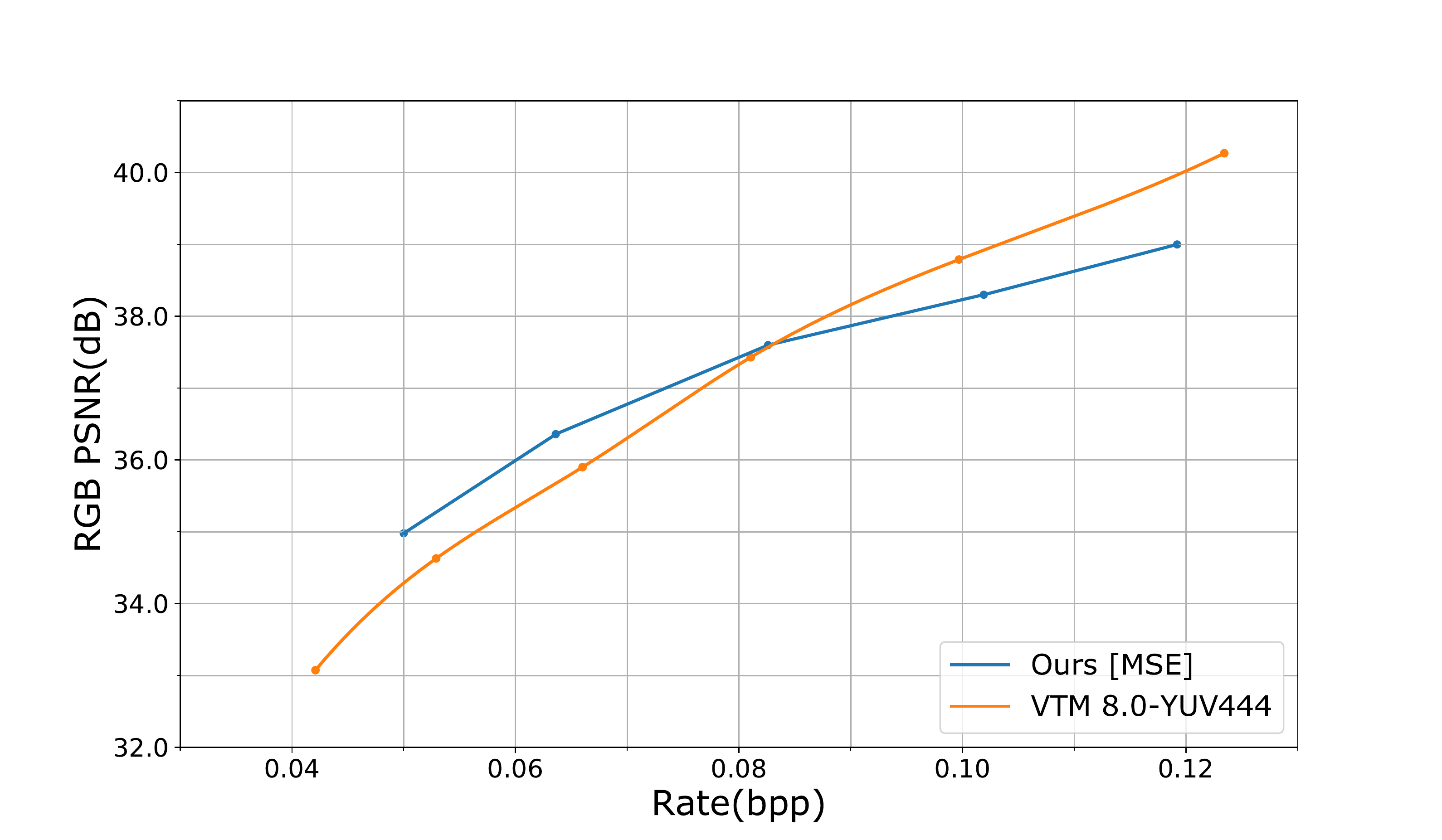}
\caption*{Image 1}
    \end{subfigure}
    \end{minipage}
 \hspace{0.008\linewidth}
    \begin{minipage}{0.48\textwidth}
    \begin{subfigure}{\columnwidth}
\includegraphics[scale=0.31, clip, trim=1.5cm 0.3cm 2cm 2.0cm]{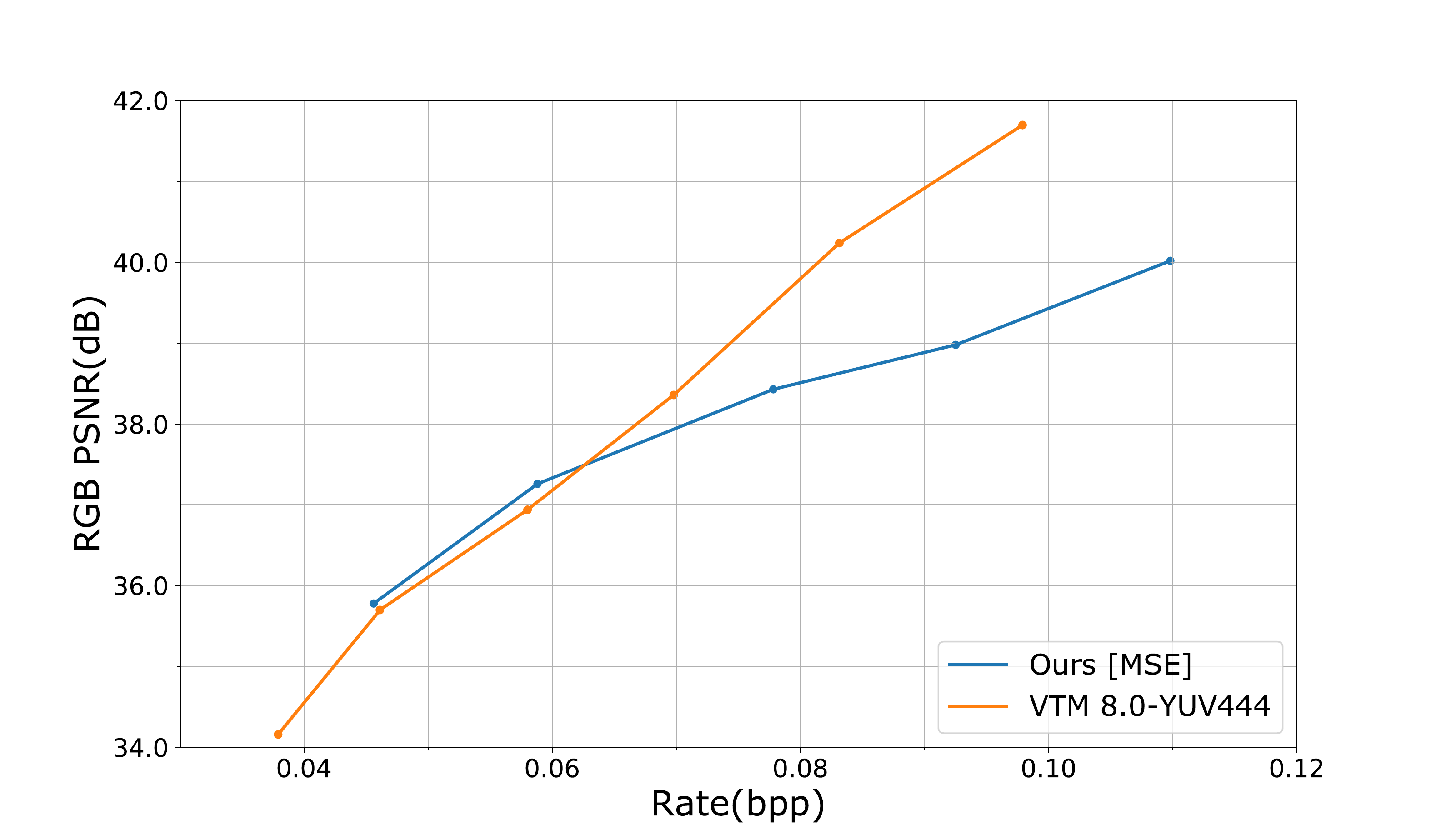}
\caption*{Image 2}
    \end{subfigure}
    \end{minipage}
\caption{Individual RD curves of two screen-captured images.}
\label{figure12}
\end{figure*}

Furthermore, our proposed method is expected to present better results on images with a lot of repeat patterns such as screen-captured images. Here, we evaluate the performance of our method on some screen-captured images from HEVC standard test sequences of screen content. We notice that there are some screen-captured images with very simple contexts, which require only few bits ($\textless$ 0.01 bpp) but would easily deliver a high-quality reconstruction (PSNR $\textgreater$ 45dB). Therefore, this kind of images would have significant influence on the average rate-distortion performance and makes the average statistics unreliable. We instead provide individual RD curves of two specific screen-captured images to compare our method with VVC. As shown in Fig. \ref{figure12}, we use our full compression model that is still trained on the ImageNet dataset for comparison. It can be observed that our method achieves better performance when the bitrate is low. It verifies the effectiveness of our proposed causal prediction method. However, VVC gradually performs better than our method when the bitrate and PSNR value increase. This result can be explained from the perspective of autoencoder (AE) limit of neural compression model \cite{helminger2020lossy}. Our neural compression model is a VAE-based model, where the encoder transform and the decoder transform are non-invertible. It is different from the linear invertible transform such as DCT used in traditional codecs. The non-invertible transform network introduces errors and information loss, thereby sets a lower bound of distortion. In contrast, invertible transform used in traditional codecs does not bring in information loss. When the reconstruction quality is very high, the learning-based compression model would reach the AE limit and performs worse than traditional codecs. Since screen-captured images have a lot of repeated patterns and some even have very simple contexts, the reconstruction quality of screen-captured image is naturally very high. Therefore, when using learning-based codecs to compress screen-captured images, it will easily encounter the AE-limit issue due to the high-quality reconstructions. As a result, in this case of compression of screen-captured images, our method performs better than VVC when the bitrate is low but is not excellent as VVC when the PSNR value is high..

\subsection{Ablation Study}

In Section \ref{section4d}, we conduct an exploratory experiment to investigate the effects of different prediction modes where the baseline is a relative simple model \cite{minnen2018joint}. To further verify the effectiveness of different modules, we conduct ablation studies under the same experimental settings where the baseline is now our main compression network.

\begin{figure}[t]
\centering
     \includegraphics[scale=0.31, clip, trim=1.3cm 0cm 1cm 1.6cm]{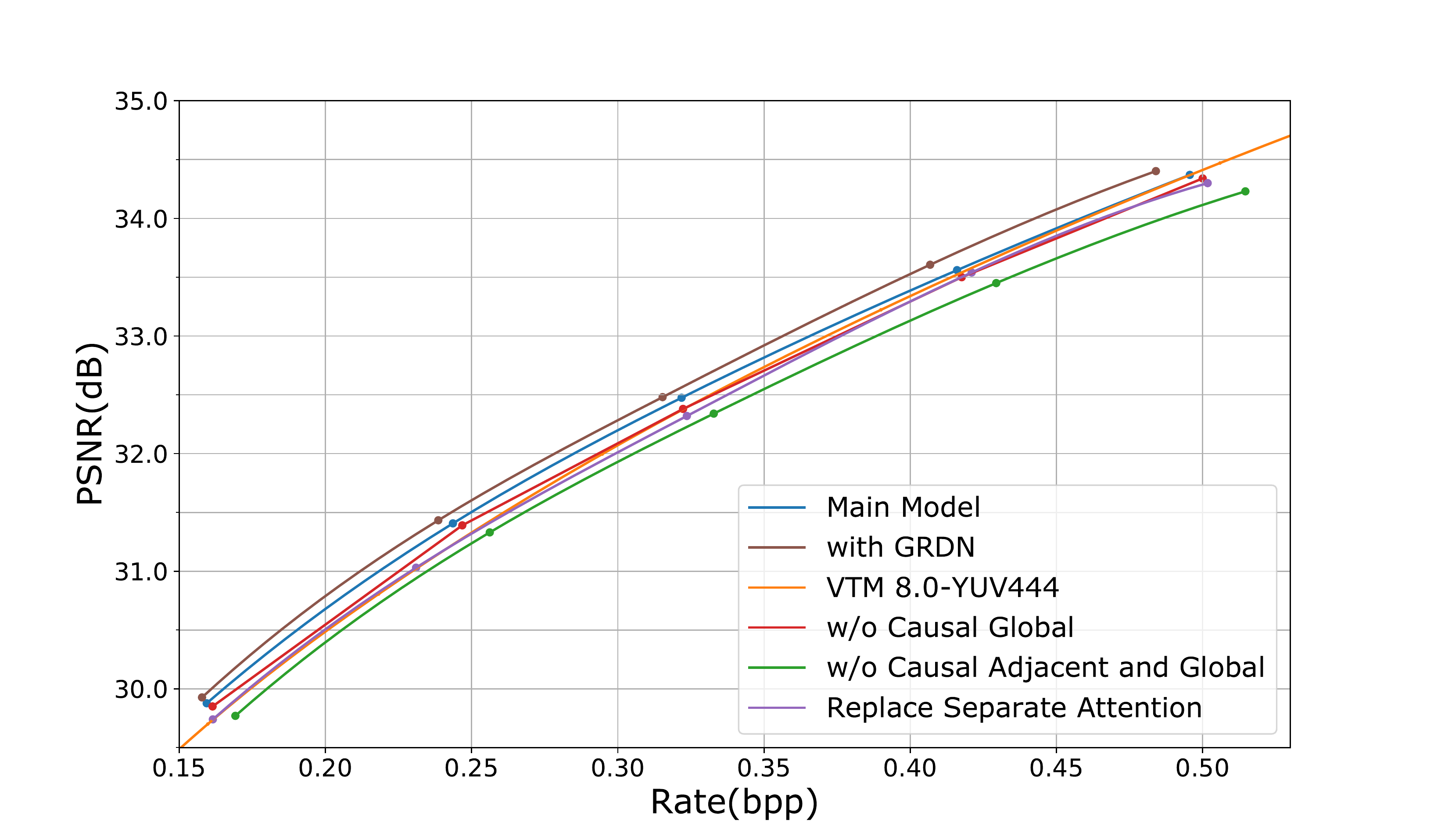}
\caption{Ablation study where baseline is our main compression model.}
\label{figure13}
\end{figure}

As shown in Fig.\ref{figure13}, \textit{Main Model} represents our main compression network without postprocessing. Additional postprocessing GRDN boosts performance by approximately 0.1 dB (the brown line) and this corresponds to the optimal performance of our method in Fig.\ref{figure10}. Compared with the main compression network, both the causal adjacent context model and causal global prediction model are important to achieve improved rate-distortion performance (as shown by the red line and the green line, respectively). Note that \textit{w/o causal global} refers to removing the causal global prediction model while preserving causal context model. Considering that our baseline network is powerful enough, such improvements are encouraging. 
In addition, replacing the proposed separate attention module by the attention layer used in \cite{cheng2020learned} also results in a performance drop. In conclusion, our proposed three elements, including causal context model, causal global prediction model and separate attention layer, contribute to the state-of-the-art performance of our method.

\begin{table}[t]
\centering
\small
\resizebox{\columnwidth}{!}{
\renewcommand{\arraystretch}{1.2}
\begin{tabular}{|c|c|c|}
    \hline
      & Encoding Time (s) & Decoding Time (s) \\
    \hline
	Masked Context \cite{minnen2018joint} & 3.4 & 6.7 \\
    \hline
     Causal Context & 3.5 & 7.9 \\
    \hline
     \makecell{Causal Context + \\ Causal Global Prediction} & 3.5 & 38.7 \\
    \hline
\end{tabular}}
\caption{Encoding and decoding time on Kodak dataset. The statistics are averaged on the whole dataset. Those three methods have the similar encoding time because encoding can be computed in parallel.}
\label{table2}
\end{table}
\subsection{Coding time}

To evaluate the running time of the models fairly, we limit the accessible resources to one 2080 Ti GPU and two cores of an Intel E5-2699 v4 CPU. 
We modify our previous implementation submitted to CLIC 2020 to evaluate the coding time of three different entropy models. Here, \textit{Mask Context} is the conventional context model with 2-D mask convolution \cite{minnen2018joint} that is taken as the base model. \textit{Causal Context} corresponds to our submission to CLIC competition \cite{guo20203} which adopts the adjacent causal context model. As shown in Table II, since the serial decoding separates the decoding process across channels, the decoding time increases from average 6.7 seconds to average 7.9 seconds. Note that we test all the 24 Kodak images of which resolutions are 768$\times$512. \textit{Causal Context + Causal Global Prediction} is the most powerful entropy model proposed in this paper. However, despite the state-of-the-art rate-distortion performance, this complex entropy model requires 37.7 seconds for decoding on average. It is unsurprising because the global searching of references points consumes much more time. There remains much room for optimization in terms of both software and hardware.



\section{Conclusion\label{section6}}

In this paper, we explore reducing the global redundancies and cross-channel redundancies among the latent variables in an entropy model. It is observed that separating the latents is advantageous for improving the entropy model.
To this end, we first propose a causal context model to generate highly informative adjacent context. We then extend it to a causal global prediction model to conduct global prediction with accurate global references. Both models make use of channel-wise redundancies to facilitate entropy estimation for a specific latent group. While the separate entropy model suffers from an increased decoding time, it significantly improves the rate-distortion performance. In addition, we adopt a new group-separated attention module that enables independent feature-map attention in separate groups, thereby enhancing the transform networks. Experimental results indicate that our method achieves the state-of-the-art performance in terms of both PSNR and MS-SSIM.

\bibliographystyle{paper}
\bibliography{references}

\begin{IEEEbiography}[{\includegraphics[width=1in,height=1.25in,clip,keepaspectratio]{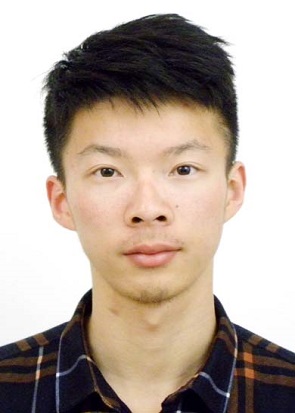}}]{Zongyu Guo}
	received the B.S. degree from University of Science and Technology of China in 2019. He is currently a Ph.D. student in the Department of Electronic Engineering and Information Science, in University of Science and Technology of China, advised by Prof. Zhibo Chen. His research interests include image/video compression, image inpainting, and generative modelling of data distribution.
\end{IEEEbiography}

\begin{IEEEbiography}[{\includegraphics[width=1in,height=1.25in,clip,keepaspectratio]{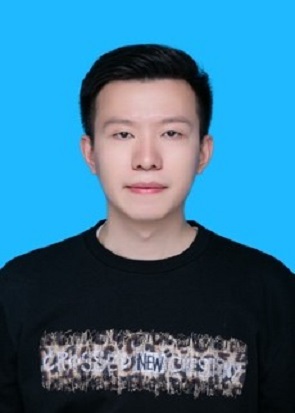}}]{Zhizheng Zhang}
	received the B.S. degree in electronic information engineering from the University of Electronic Science and Technology of China, Chengdu, China, in 2016. He is currently pursuing the Ph.D. degree with the University of Science and Technology of China, Hefei, China. His current research interests include image/video compression, intelligent media understandings, and reinforcement learning.
\end{IEEEbiography}

\begin{IEEEbiography}[{\includegraphics[width=1in,height=1.25in,clip,keepaspectratio]{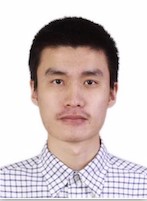}}]{Runsen Feng}
	received the B.S. degree in electronic information engineering from the University of Science and Technology of China, Hefei, China, in 2018. He is currently a Ph.D. student at the University of Science and Technology of China. His research interests include learning-based image coding and video coding.
\end{IEEEbiography}

\begin{IEEEbiography}[{\includegraphics[width=1in,height=1.25in,clip,keepaspectratio]{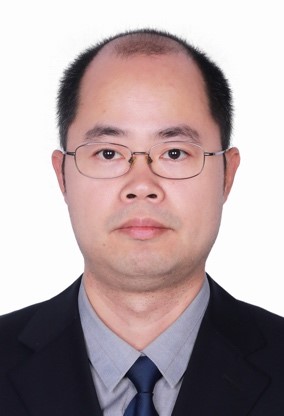}}]{Zhibo Chen}
	(M'01-SM'11) received the B. Sc., and Ph.D. degree from Department of Electrical Engineering Tsinghua University in 1998 and 2003, respectively. He is now a professor in University of Science and Technology of China. His research interests include image and video compression, visual quality of experience assessment, immersive media computing and intelligent media computing. He has more than 150 publications and more than 50 granted EU and US patent applications. He is IEEE senior member, Secretary/Chair-Elect of IEEE Visual Signal Processing and Communications Committee, and member of IEEE Multimedia System and Applications Committee. He was TPC chair of IEEE PCS 2019 and organization committee member of ICIP 2017 and ICME 2013, served as TPC member in IEEE ISCAS and IEEE VCIP.
\end{IEEEbiography}

\end{document}